\documentclass[10pt,journal,compsoc]{IEEEtran}
\ifCLASSOPTIONcompsoc
  \usepackage[nocompress]{cite}
\else
  \usepackage{cite}
\fi
\ifCLASSINFOpdf
\usepackage{graphicx}
\usepackage{subcaption}
\usepackage{tikz}
\usepackage{comment}
\usepackage{amsmath,amssymb} 
\usepackage{color}
\usepackage{multirow}
\else
\fi
\begin{document}
%
\title{CP3: Unifying Point Cloud Completion by Pretrain-Prompt-Predict Paradigm}
%
%
%
%
\author{Mingye Xu,
	Yali Wang,
	Yihao Liu,
	Tong He,
	Yu Qiao, ~\IEEEmembership{Senior Member, IEEE}
\IEEEcompsocitemizethanks{\IEEEcompsocthanksitem 
	M. Xu, Y. Wang and Y.  Liu are with  Shenzhen Institute of	Advanced Technology, Chinese Academy of Sciences, Shenzhen, 518055. Email: my.xu@siat.ac.cn.
\IEEEcompsocthanksitem M. Xu and Y. Liu are also with the University of Chinese Academy of Sciences, Beijing,
100049. 
 Y. Wang, Tong He and Y.Qiao are  with Shanghai AI Laboratory.
\IEEEcompsocthanksitem Y. Wang and Y.Qiao are co-corresponding authors.}
}

%
%

\markboth{ }%
{Shell \MakeLowercase{\textit{et al.}}: Bare Demo of IEEEtran.cls for Computer Society Journals}
%



\IEEEtitleabstractindextext{%
\begin{abstract}
Point cloud completion aims to predict complete shape  from its partial observation. Current approaches mainly consist of generation and refinement stages in a coarse-to-fine style.
However, the generation stage often lacks robustness to tackle different incomplete variations, while the refinement stage blindly recovers point clouds without the semantic awareness.
To tackle these challenges,
we unify point cloud \textbf{C}ompletion by a generic \textbf{P}retrain-\textbf{P}rompt-\textbf{P}redict paradigm,
namely \textbf{CP3}.
Inspired by  prompting approaches from NLP, we creatively reinterpret point cloud generation and refinement as the prompting and predicting stages, respectively. Then, we introduce a concise self-supervised pretraining stage before prompting. It can effectively increase robustness of point cloud generation, by an Incompletion-Of-Incompletion (IOI) pretext task.
Moreover, we develop a novel Semantic Conditional Refinement (SCR) network at the predicting stage. 
It can discriminatively modulate multi-scale  refinement with the guidance of semantics. 
Finally, extensive experiments demonstrate that our CP3 outperforms the state-of-the-art methods with a large margin.

\end{abstract}

}

\maketitle

\IEEEdisplaynontitleabstractindextext

%
\IEEEpeerreviewmaketitle

\IEEEraisesectionheading{\section{Introduction}\label{sec:intro}}


Point cloud analysis has attracted a lot of research interest in computer vision and robotics.
Unfortunately,
the scanned point cloud is usually incomplete in practice,
due to complex occlusion, light reflection, limited sensor resolution, etc.
Therefore,
the completion task has gradually become important.
The existing approaches \cite{yuan2018pcn,wang2020cascaded,liu2020morphing,pan2020ecg,pan2021variational}
are mainly based on point cloud generation and refinement,
which first generates the coarse completion from partial point cloud,
and then recovers the fine completion from the coarse one.
However,
these approaches focus on designing complicated models,
while ignoring the fundamental problems about how to effectively use point cloud at different stages.

\begin{figure*}[t]
	\centering
	\includegraphics[width=0.97\linewidth]{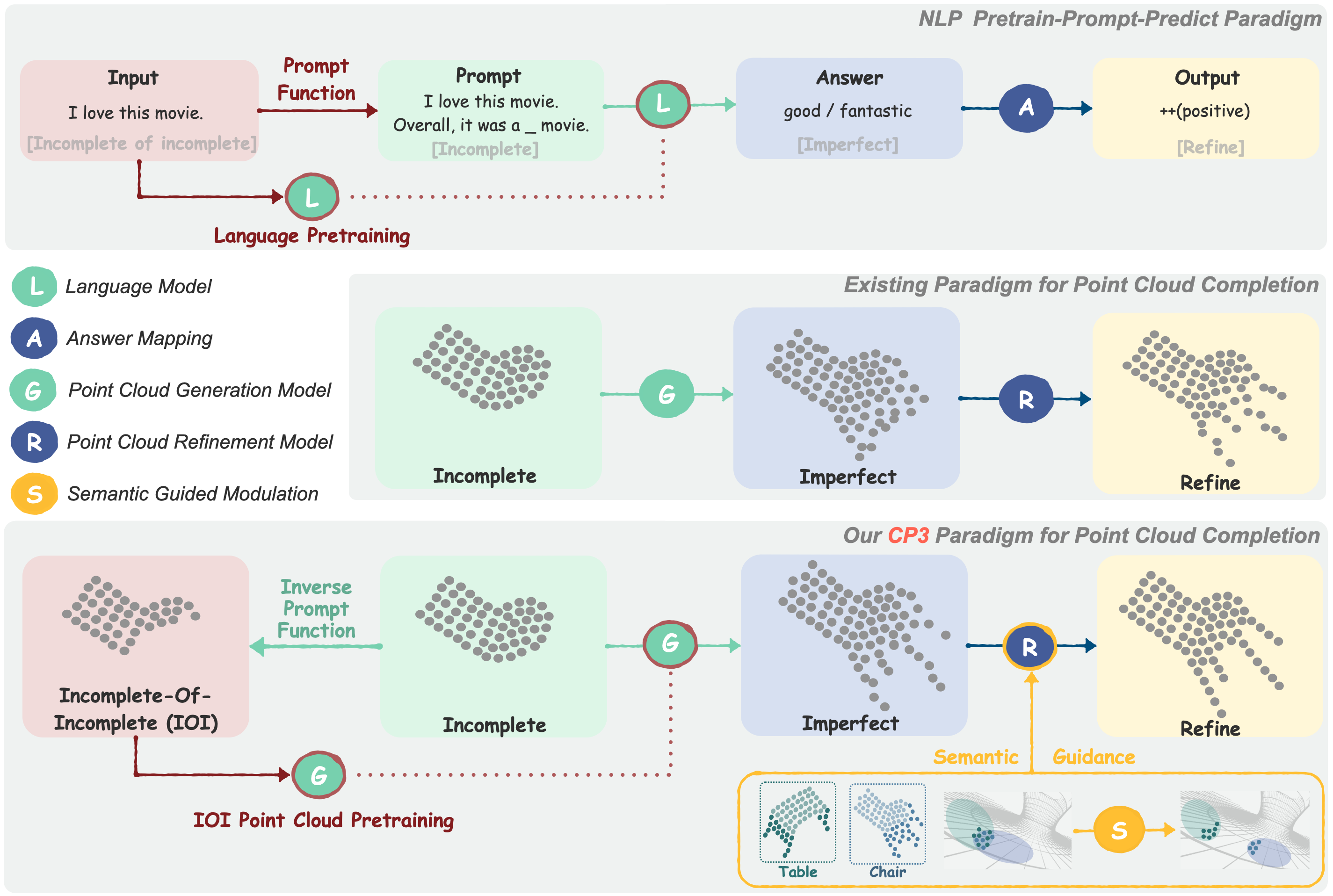}
	\caption{
		Our CP3 paradigm.
		We unify point cloud completion by the recent prompting scheme in NLP,
		since we find the \textit{incomplete} point cloud plays a similar role as the \textit{incomplete} sentence,
		e.g., 
		``I love this movie. Overall, it was a $\_\_$ movie".
		Following this line,
		we reinterpret point cloud generation and refinement as the prompting and predicting stages,
		which first generates the imperfect point cloud (like \textit{answer}: ``good / fantastic" ) from the incomplete one,
		and then refines target point cloud (like \textit{output}: ``++(positive)" ) from the imperfect one.
		But different from NLP,
		point cloud generation is not pretrained,
		and refinement lacks semantic awareness.
		Hence,
		we propose a concise IOI pretraining to boost generation robustness in a self-supervised way,
		and design a novel semantic guidance to modulate refinement for shape confusion reduction.}
	\label{fig_intro_1}
\end{figure*}

At the generation stage,
the core problem is to learn a robust incomplete-to-complete mapping for tackling different partial observations of point cloud.
Most methods mainly work on developing deterministic or probabilistic generation networks \cite{wang2020cascaded,pan2020ecg,pan2021variational},
while lacking insightful consideration about data itself.
In fact,
the training pairs of incomplete and complete point cloud are predetermined.
With such limited data diversity,
it is often difficult to generate robust completion from new incomplete variations,
by training those complex generation networks from scratch.
At the refinement stage,
the key is to recover the plausible shape of the complete point cloud,
based on the coarse (or imperfect) completion from the generation stage.
However,
the existing refinement approaches \cite{xie2020grnet,xiang2021snowflakenet,pan2020ecg} ignore the semantic information of point cloud,
which is an important clue of shape recovery. 
As a result,
these refinement networks are unsatisfactory on point clouds which have similar shapes but belong to different semantic categories,
e.g.,
chair and table in Figure \ref{fig_intro_1}.

To tackle these problems,
we propose to unify point cloud \textbf{C}ompletion frameworks by a generic \textbf{P}retrain-\textbf{P}rompt-\textbf{P}redict paradigm,
namely \textbf{CP3},
which can achieve robust generation and discriminative refinement by self-supervised pretraining and semantic-guided predicting.
As shown in Figure \ref{fig_intro_1},
our inspiration mainly comes from the recent prompting scheme in NLP \cite{liu2021pre,wei2021finetuned,brown2020language}. 
Specifically,
as Figure \ref{fig_intro_1} shows, 
we flexibly reinterpret the generation stage as the prompting stage,
since both stages aim at learning on new scenarios with limited data diversity.
But differently,
the pretrained model is ready in NLP.
Therefore,
its goal is to design a \textit{prompt function} that modifies $input$ as $prompt$ for answer search.
Alternatively,
point cloud generation is not pretrained.
Hence,
our goal is to design an \textit{inverse prompt function} which aims at
constructing the diversified $input$ from the given $prompt$ (i.e., incomplete point cloud) for pretraining.
In particular,
we introduce a concise Incompletion-Of-Incompletion (IOI) sampling mechanism as \textit{inverse prompt function},
where
we randomly crop incomplete point clouds into new incomplete-of-incomplete ones.
Subsequently,
we pretrain the generation network with a self-supervised pretext task,
which 
applies the IOI point cloud as input to recover the original incomplete point cloud. 
This design contains two main advantages.
First,
this pretraining task is similar to the downstream generation task,
which can reduce the task gap for model transfer \cite{liu2021pre}. 
Second,
IOI pretraining can effectively increase robustness of generation network,
by pretraining on rich variations of incomplete point clouds.
Specially, 
our  IOI pretraining is not affected by the fixed form of crop procedure,
we verify  different ways of crop procedure (as indicated in the Figure \ref{fig_IOIaugment} ),
 both with improved performance significantly.

Moreover,
we reinterpret the refinement stage as the predicting stage,
since this stage is used to map the imperfect point cloud (i.e., $answer$ of $prompt$) into the refined one (i.e., the target $output$).
But different from answer mapping in NLP,
point cloud refinement requires extra semantic knowledge,
since 
the shape of point cloud is highly relevant to its high-level semantics, such as  category information.
Based on this observation,
we propose a novel Semantic Conditional Refinement (SCR) network,
which  adaptively leverage semantic information as discriminative guidance for refining point cloud.
Specifically,
it mainly consists of two distinct blocks,
i.e.,
semantic-guided modulation and multi-scale point deconvolution blocks.
The semantic-guided modulation block smartly converts the semantic information of point cloud into a dynamical filter to modulate the point-wise features,
or  creates the  affine transformations with the semantic information for feature modulation.
Conditioned on these semantic-modulated features,
the point deconvolution block progressively refines point cloud by multi-scale relation learning.
Hence,
unlike the previous refinement networks \cite{pan2021variational,xiang2021snowflakenet,yu2021pointr},
our SCR network can effectively alleviate shape confusion among various categories and semantics.

Finally, 
we evaluate our CP3 on the widely-used point cloud completion benchmarks
such as MVP \cite{pan2021variational} and PCN \cite{yuan2018pcn}.
The extensive experiments show that our IOI pretraining and SCR predicting can effectively boost point cloud generation and refinement.
Consequently,
CP3 outperforms a number of state-of-the-art approaches with a large margin,
e.g.,
it achieves the new SOTA result on both MVP and PCN datasets,
with 2.27 CD loss on MVP (the current SOTA \cite{pan2021variational} is 3.06) and 7.02 CD loss on PCN (the current SOTA \cite{xiang2021snowflakenet} is 7.21).
We will release our code afterwards.

\section{Related Work
	\label{sec:relatedwork}}

\subsection{3D shape Completion}
Most traditional completion methods  rely on shape continuity inferred from input scan.
Some methods \cite{shalom2010cone,attene2010lightweight,lin2010fusion,hu2019local,nguyen2016field} are mainly based on hand-crafted rules or smooth interpolation/extrapolation to characterize the missing regions.
Generative based methods \cite{shen2012structure,shao2012interactive,martinovic2013bayesian} is to  exploit the large-scale shape database to search for the similar shapes/patches to fill the missing regions of 3D shapes. 
However, the generalization capability of these methods to diverse  partial structures  is usually limited.
In contrast,  deep learning methods can learn more flexible representations  for  more complicated shapes. 
The existing deep learning  methods  are mainly based on the crose-to-fine style, which generates a coarse one and refines the geometry details.
PCN \cite{yuan2018pcn}  first operates on raw point clouds and  generates a coarse completion based on learned global representation and folding based decoding \cite{yang2018foldingnet}.
Based on PCN, many methods  \cite{yuan2018pcn,wang2020cascaded,liu2020morphing,pan2020ecg}
achieve more detailed point cloud completion with better completion results,
but usually lack the long-range correlations in local regions.
\cite{pan2021variational,yu2021pointr,xiang2021snowflakenet} adopt transformers \cite{vaswani2017attention} to learn the structural information of pairwise interactions and context correlations.
However, they pay much attention to the complicated model structures and usually integrate the generation designs and refinement designs in a single framework, while ignoring deep investigation of distinct data problems in each stage.
In addition,
a self-supervised approach has been introduced for scene completion with RGB-D scans \cite{dai2020sg},
while different data formulation and task definition make it hard for traditional point cloud completion.
Different from the previous approaches,
we creatively reinterpret the existing completion process as prompt (generation) and predict (refinement),
inspired by the ``pretrain-prompt-predict" paradigm in NLP \cite{liu2021pre}.
Moreover,
we propose an IOI pretraining for data-diversified generation, 
and develop a discriminative refinement network with semantic guidance.




\subsection{Conditional Feature Modulation}

Conditional feature modulation has been used extensively in prior and concurrent works including 
discrete labels \cite{mirza2014conditional,goodfellow2014generative}, 
text \cite{reed2016generative},
and images \cite{isola2017image}. 
For example,
Condition GAN \cite{mirza2014conditional} can generate the controllable images conditioned on class labels.
Specifically, adding constraints on the conditional information in the generator and discriminator  makes GAN training easier and more stable.
AdaIN \cite{huang2017arbitrary} can effectively achieve image transfer by matching feature statistics of content image and style image and  retaining the  structure of the content image.
Inspired by style transfer task,
StyleGANs \cite{karras2019style,karras2020analyzing} design modulations in the generator to adjust the `style' of an image, 
 thus directly control the   mage generation process.
Since different semantic priors have different effects on different regions of the image,
\cite{he2020conditional} propose a global feature modulation method for photo retouching.
\cite{wang2018recovering} propose a spatial feature-based modulation method to generate super-resolution images with more natural and realistic texture images.
Inspired by these modulation methods,
we introduce the semantic-guided modulation for point cloud completion,
which can effectively reduce shape confusion for discriminative refinement.
Moreover, unlike PointNet++ \cite{qi2017pointnet++} for part segmentation, which concatenates semantic information directly onto point-wise feature representation, our semantic-guided modulations are more efficient and lightweight.



\subsection{Prompting Paradigm}
Recently, 
the prompting paradigm has attracted a lot of attention in NLP \cite{liu2021pre,lester2021power,lu2021fantastically}.
The conventional training paradigm refers to pretrain-finetune,
while the prompting paradigm is characterized as pretrain-prompt-predict.
Specifically,
it is used to rebuild the input text through some prompt information, 
and transfer the predetermined task to the prompting task,
e.g., 
``I love this movie." is changed into ``I love this movie. Overall, it was a  \_\_ movie.".
 Unlike supervised learning, 
 prompting dose not need to fine tune the whole model, but to design a text prompt to rebuild the model.
 Specifically, it transfers the knowledge in a large number of pretrained models and rearranges prompting tasks to make them  closer to pre-determined tasks.
By designing appropriate prompts,
a number of works \cite{radford2019language,brown2020language,petroni2019language}  generalizes the model behavior well,
so that the pretrained language model can be effectively adapted to predict the desired output in NLP.
Moreover,
such prompting paradigm has been also introduced for vision-language modeling in computer vision \cite{zhou2021learning}.
To our best knowledge,
we are the first to leverage such ``Pretrain-Prompt-Predict" paradigm for point cloud completion,
where we unify point cloud generation and refinement as the prompting and predicting stage, 
and introduce a self-supervised pretraining stage with the help of our inverse prompt function.

\begin{figure*}[t]
	\centering
	\includegraphics[width=0.85\linewidth]{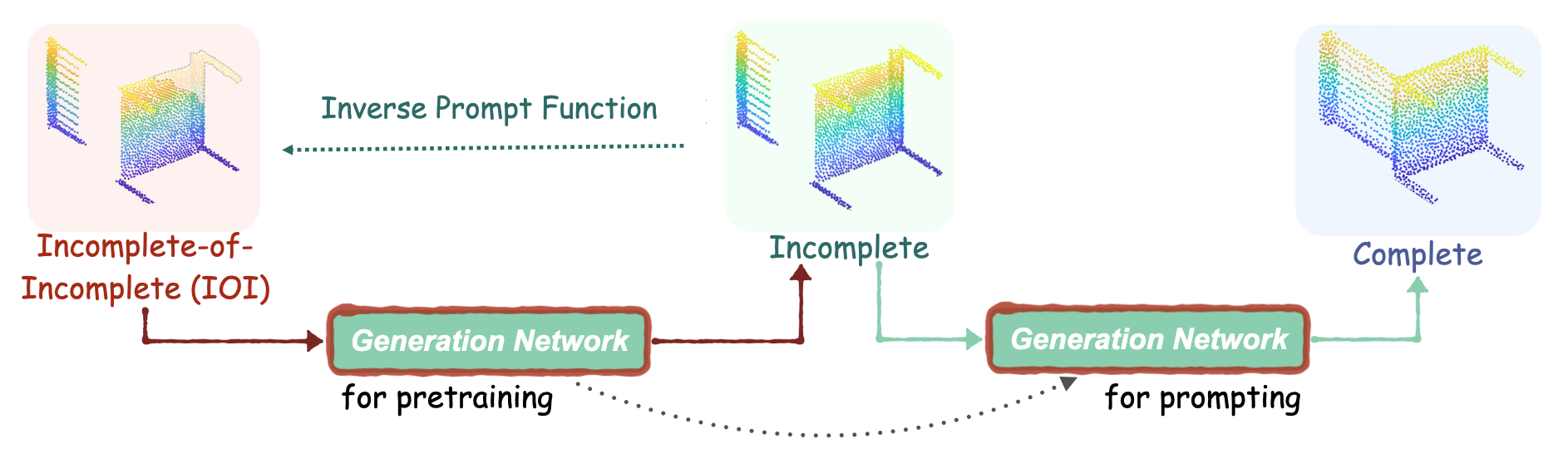}
	\caption{ IOI pretraining and prompting. 
	}
	\label{fig_IOIpretrain_prompt}
\end{figure*} 

\section{Method
	\label{sec:method}
}

\begin{figure}[t]
	\centering
	\includegraphics[width=1\linewidth]{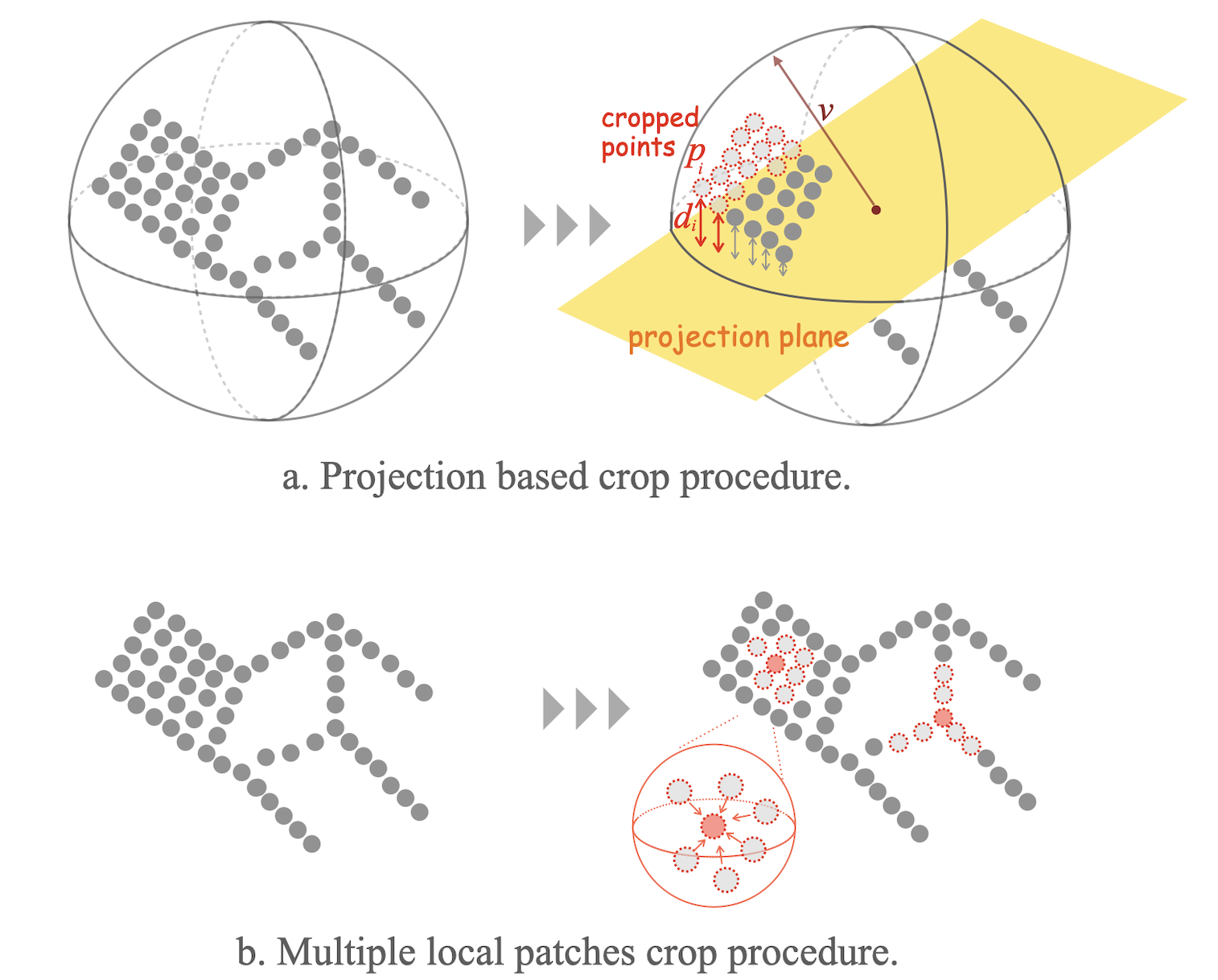}
	\caption{Two types of crop procedure of IOI sampling.
	}
	\label{fig_IOIaugment}
\end{figure}

\textbf{Overall Paradigm of Our CP3}.
In this section,
we introduce the proposed CP3 paradigm in details.
As shown in Figure \ref{fig_intro_1},
we unify point cloud completion by the recent prompting paradigm in NLP.
Specifically,
point cloud generation refers to the prompting stage.
Given an incomplete point cloud (i.e., \textit{prompt}),
the generation network is to produce an imperfect completion (i.e., \textit{answer}).
Since the training pairs in this stage are predefined with limited data diversity,
learning the generation network from scratch leads to unsatisfactory overfitting.
To tackle the above problem,
we introduce a pretraining stage for point cloud generation.
In particular,
we design an \textit{inverse prompt function},
i.e.,
Incompletion-Of-Incompletion (IOI) sampling,
which randomly crops incomplete point clouds (i.e., \textit{prompt}) into a large number of diversified IOI point clouds (i.e., \textit{input}).
Subsequently,
we pretrain the generation network with a self-supervised pretext task,
where
we use IOI point cloud as input to reconstruct the corresponding incomplete one.
Finally,
we reinterpret the refinement stage as the predicting stage,
where
we predict the target point cloud (i.e., \textit{output}) from the imperfect or coarse one (i.e. \textit{answer} in the prompting stage).
To enhance such an answer mapping,
we introduce a novel Semantic Conditional Refinement (SCR) network.
It can effectively reduce shape confusion by progressively refining multi-scale point cloud with its semantic category.

\subsection{\textit{Pretrain} \& \textit{Prompt}: Point Cloud Generation with IOI-Pretrain}

As discussed before,
we rethink point cloud generation by the prompting stage in NLP,
where 
an incomplete point cloud is analogous to
an incomplete sentence,
e.g.,
``I love this movie. Overall, it  was a $\_\_$ moive''.
But differently,
a well-pretrained language model is ready for the prompting stage in NLP,
while
such a model is not available for point cloud generation.
In fact,
the existing generation models \cite{pan2021variational,pan2020ecg,yuan2018pcn} are trained from scratch,
with limited diversity of training pairs.
As a result,
its performance would be restricted,
when tackling new incomplete variations.
To enhance its robustness,
we propose a self-supervised pretraining stage for the generation model,
based on a concise Incompletion-Of-Incompletion (IOI) sampling mechanism.

\quad

\noindent	\textbf{IOI Sampling as Inverse Prompt Function.}
To construct the diversified inputs for pretraining,
we propose two types of concise IOI sampling mechanisms,
which generate incomplete-of-incomplete point clouds (i.e., \textit{input}) from incomplete ones (i.e., \textit{prompt}).
We call IOI as \textit{inverse prompt function},
with inspiration of \textit{prompt function} in NLP.
The difference is that,
the prompt function in NLP is to convert \textit{input} as \textit{prompt} for answer search,
while 
our inverse prompt function is to construct diversified \textit{input} from the given \textit{prompt}.
Specifically,
we offer two optional crop procedures for IOI sampling in Figure \ref{fig_IOIaugment}:

\textbf{Option a:} Projection based crop procedure.
{First}, 
we randomly determine an projection plane through the given incomplete point cloud. 
In particular, a random vector $v$ is ascertained from the object center through two random angle $\phi$ and $\theta$, where $ {v} =\{ {\sin(\theta) \cos(\phi)}, \sin(\theta) \sin(\phi),  \cos(\theta) \}$. 
Then we find  the projection plane which is perpendicular to this vector $v$ and passes through the center of point cloud
(i.e., yellow plane in Figure \ref{fig_IOIaugment}):
{Second}, 
we calculate the projection distance ${d}_i $ from each point ${p}_i $ to this projection plane, 
\begin{equation}
{d}_i = ({p}_i -\frac{1}{N}\sum\nolimits_{j=1}^{N}p_{j})\circ {v},
\label{eq:distance}
\end{equation}
where $\circ $ is the hadamard product.
{Finally}, 
we use the projection distance ${d}_i$ as sampling guidance,
and drop the farthest $r N $ points from the projection panel,
i.e., red points in Figure  \ref{fig_IOIaugment}.
The crop rate is set as $r \subset [0, \Gamma ]$, 
where
$\Gamma$ is the maximun threshold.

\begin{figure*}[t]
	\centering
	\includegraphics[width=1\linewidth]{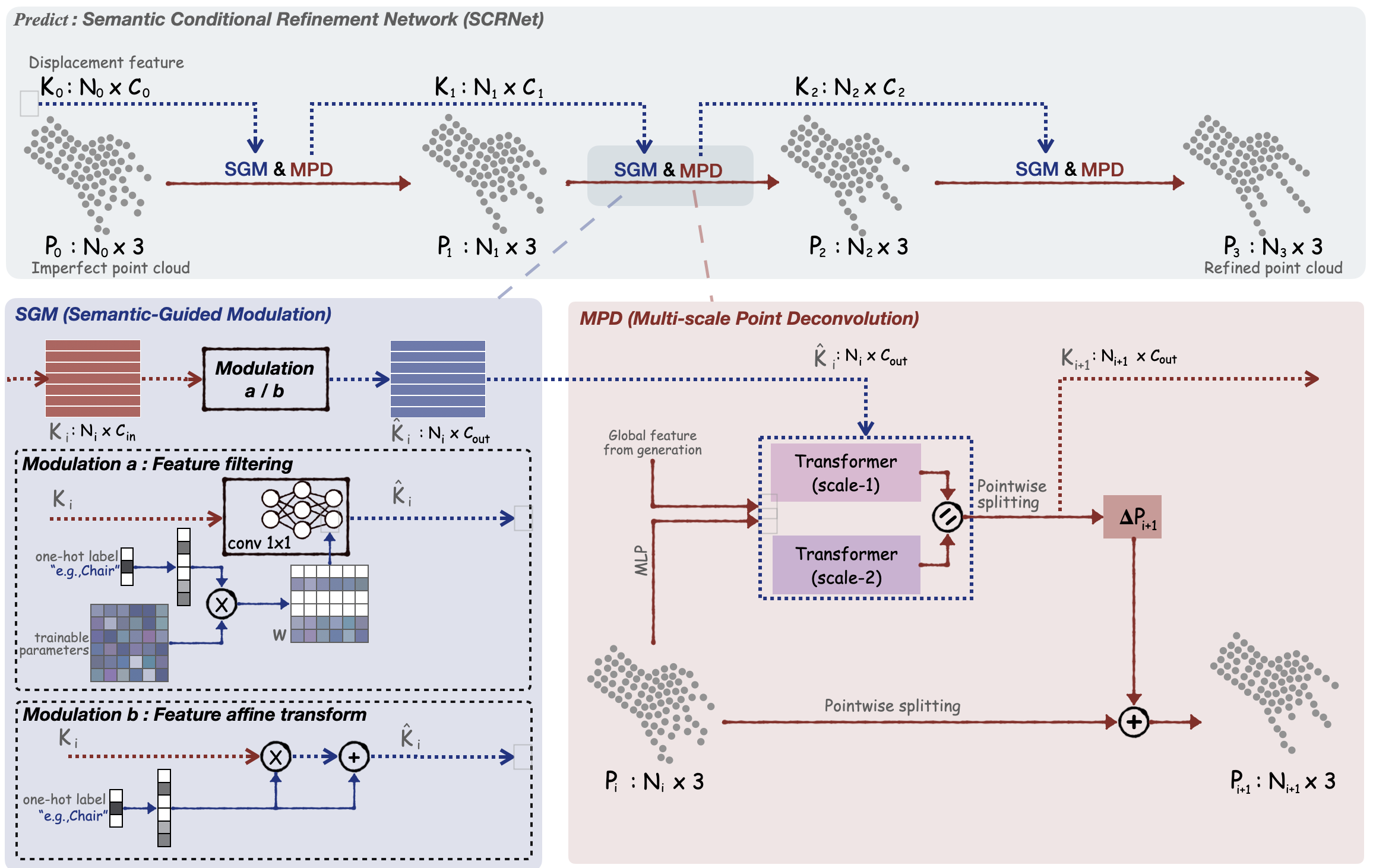}
	\caption{The structure of SCR network with Semantic-Guided Modulation and Multi-scale Point Deconvolution. }
	\label{prnet}
\end{figure*}

\textbf{Option b:} Multiple local patches crop procedure.
We can also  take the  common used mask mechanism \cite{pang2022masked} on point cloud for IOI sampling,
where we randomly crop $\mathbf{\textit{m}}$  local patches in the point cloud.
First, we use furthest point sampling method to sample $\mathbf{\textit{m}}$ points as the centers of cropped patches.
Then we   select the nearest points around the $\mathbf{\textit{m}}$ centroids as the crop points, e.g., red points in Figure  \ref{fig_IOIaugment}.
In contrast, multiple local patches  crop procedure results in more random incomplete structures of the point cloud.

It is worth mentioning that,
both projection based crop  and multiple local patches crop  of IOI sampling can  bring the following benefits.
On one hand,
the random crop of incomplete point cloud can generate rich data diversity for pretraining.
On the other hand,
it naturally builds up a pretext task for self-supervised pretraining as follows.


\quad

\noindent	\textbf{Self-Supervised IOI Pretraining}.
By IOI sampling,
we have incomplete point clouds and their incomplete-of-incomplete variations on hand.
Hence,
we can pretrain the generation network by self-supervised learning of incomplete point cloud,
i.e.,
we recover the incomplete point cloud from its IOI point cloud,
as shown in Figure \ref{fig_IOIpretrain_prompt}.
Moreover,
we would like to emphasize that,
our pretraining task is actually similar to the original generation task (i.e., the prompting stage),
by analogy with the``incomplete-to-complete'' mapping.
Hence,
such design can effectively reduce the task gap for model transfer,
which perfectly matches the insight of prompting in NLP \cite{liu2021pre,lester2021power}.
Finally,
we choose the recent VRCNet \cite{pan2021variational} as an exemplar of the generation network for pretraining.
Note that,
other generation networks also work with our pretraining design.
More details can be found in Subsection \ref{sec:ablation} and in Table \ref{tab:ablation_IOI_other_methods}.

\quad

\noindent	\textbf{Prompting}.
At the generation stage,
we are actually given with limited training pairs of incomplete and complete point clouds.
Hence,
we simply fine-tune the generation network with these pairs,
after pretraining.
Consequently,
we use the well-tuned network for prompting,
i.e.,
generating a coarse complete one (i.e., \textit{answer}) from an  incomplete point cloud (i.e., \textit{prompt}).

\subsection{\textit{Predict}: Point Cloud Refinement with Semantic Guidance}

After obtaining the coarse point cloud (i.e., \textit{answer}) from the incomplete one (i.e., \textit{prompt}),
we next map \textit{answer} into \textit{output},
i.e.,
the target complete point cloud.
As discussed in the introduction,
this predicting stage refers to point cloud refinement.
However,
the existing refinement networks \cite{pan2020ecg,xie2020grnet} ignore semantic understanding when recovering the shape of point cloud.
This leads to unsatisfactory completion on those confused categories.
To tackle this problem,
we propose a novel Semantic Conditional Refinement Network (SCRNet),
which progressively modulates multi-scale point cloud representation,
with guidance of high-level semantics.
As shown in Figure \ref{prnet},
it mainly consists of two core blocks,
i.e., 
Semantic-Guided Modulation (SGM) block
and  
Multi-scale Point Deconvolution (MPD) block.
In each refinement unit,
the SGM block first adjusts point-wise representation discriminatively,
by a dynamic convolution of semantic information.
Then,
the MPD block further refines the modulated representation of point cloud,
by multi-scale aggregation of geometrical context.

\quad

\noindent	\textbf{Semantic-Guided Modulation (SGM) Block}.
To reduce shape confusion,
we introduce a new Semantic-Guided Modulation (SGM) block,
which leverages semantics to module point cloud discriminatively.
We provide two effective ways of feature modulation as follows:

\textbf{Modulation a: Feature filtering.}
We mainly use the category as the semantic information for illustration.
The semantics can be also a learnable global representation of point cloud,
as shown in Table \ref{tab_ablation_modulation}.
For convenience, 
we omit the subscript $i$ for the $i$-th refinement unit in the following.
As shown in Figure \ref{prnet},
we introduce a distinct convolution with the semantic-guided filter   
$\mathbf{W} \in \mathrm{R}^{{C_{in} \times C_{out}}}$ in the SGM block.
We use this convolution to learn the semantic-relevant feature $\mathcal{\hat{K}} \in \mathrm{R}^{{N} \times C_{out}}$ from the displacement feature $\mathcal{{K}}  \in \mathrm{R}^{{N} \times C_{in}}$,
\begin{equation}
\mathcal{\hat{K}}(j,q)=\sum\nolimits_{p=1}^{C_{in}} \mathbf{W}(p,q) \cdot \mathcal{{K}}(j,p),
\end{equation}
where 
$j$ enumerates the number of points $N$, 
$p$ enumerates the number of input channels, 
and $q$ enumerates the number of output channels.
Moreover,
the displacement feature $\mathcal{{K}}$ is initialized with the point-wise feature from generation.

More specifically,
the semantic-guided filter $\mathbf{W}$ is constructed with the following steps.
First, 
we use the one-hot label vector $S$ of point cloud  to construct a category-relevant scale vector $\vartheta(S)\in \mathrm{R}^{C_{in}}$,
where
$\vartheta(.)$ refers to the MLP layer.
Second,
we use this scale vector as semantic guidance to generate the filter matrix $\mathbf{B} \in \mathrm{R}^{{C_{in} \times C_{out}}}$,
\begin{equation}
\mathbf{B}(p,q)=\vartheta(S)_{p} \cdot \mathbf{A}(p,q),
\end{equation}
where
$\mathbf{A}\in \mathrm{R}^{{C_{in} \times C_{out}}}$ is a trainable parameter matrix.
By this operation,
$\vartheta(S)_{p}$ can modulate the $p$-th  feature channel of input point cloud.
Finally,
we normalize the filter matrix for optimization stability,
\begin{equation}
\mathbf{W}(p,q)=\mathbf{B}(p,q)/ \sqrt{\sum\nolimits_{p=1}^{C_{in}} {\mathbf{B}(p,q)^{2}+\epsilon}},
\end{equation}
where $\epsilon$ is a small constant to avoid numerical issues.
Based on this normalized filter $\mathbf{W}$,
we can modulate different feature channels according to semantic category.

\textbf{Modulation b: Feature affine transform.}
Except for the feature  filtering, inspired by \cite{he2020conditional,liu2021very}, we can modulate features of another way, which is feature affine modulation.
 Feature affine transform   aims at  scaling and shifting the intermediate features $\mathcal{{K}}$ by the semantic information. 
This operation can be formulated as:
\begin{equation}
\mathcal{\hat{K}}= \sigma(\mathcal{{K}}) \odot {\alpha} + \beta
\end{equation}
 where $\odot$ denotes the element-wise multiplication operation and $\sigma$ is MLP layers,
$\mathcal{{K}}\in \mathrm{R}^{N \times C}$ is the intermediate  displacement features from MPD block, 
${\alpha}, \beta  \in \mathrm{R}^{1 \times C}$ are affine parameters that are estimated from the point cloud category labels $ S \in \mathrm{R}^{1 \times C_{cate}}$ via MLP layers.
We use conditional vector $\alpha$ to affect the cluster centers of the local representation and use conditional vector $\beta$ to fine-tune the variance in the feature space.
Thus, we can achieve point feature global adjustment with the conditional semantic information.

Through our subsequent experiments in section \ref{sec:ablation},  show that both feature filtering and affine transform can make the refinement stage more discriminative,
which can effectively distinguish point clouds that have similar local shapes but belong to different categories.
Moreover, we compare  our modulations are more effective than  other conventional modulation strategies in subsection \ref{sec:ablation}.

\quad

\begin{table*}[!t]
	\centering
	\setlength\tabcolsep{1.2pt}
	\small
	
	\begin{tabular}{l | cccccccc  cccccccc | c  }
		\hline 
		{Method} & {airp.} & {cabinet} & {car} & {chair} & {lamp} & {sofa} & {table} & {waterc.} & {bed} & {bench} & {books.} & {bus} & {guitar} & {motor.} & {pistol} & {skateb.}   & {Avg.}\\
		\hline \hline
		{ PCN} \cite{yuan2018pcn} & 	2.95 & 4.13 & 3.04 & 7.07 & 14.93 & 5.56 & 7.06 & 6.08 & 12.72 & 5.73 & 6.91 & 2.46 & 1.02 & 3.53 & 3.28 & 2.99 & 6.02 \\
		{ TopNet}\cite{tchapmi2019topnet} & 2.72 & 4.25 & 3.40 & 7.95 & 17.01 & 6.04 & 7.42 & 6.04 & 11.60 & 5.62 & 8.22 & 2.37 & 1.33 & 3.90 & 3.97 & 2.09 & 6.36 \\
		{ MSN \cite{liu2020morphing} } & 2.07 & 3.82 & 2.76 & 6.21 & 12.72 & 4.74 & 5.32 & 4.80 & 9.93 & 3.89 & 5.85 & 2.12 & 0.69 & 2.48 & 2.91 & 1.58 & 4.90 \\
		{ Wang et. al. \cite{wang2020cascaded} } & 1.59 & 3.64 & 2.60 & 5.24 & 9.02 & 4.42 & 5.45 & 4.26 & 9.56 & 3.67 & 5.34 & 2.23 & 0.79 & 2.23 & 2.86 & 2.13 & 4.30 \\
		{ ECG \cite{pan2020ecg} } & 1.41 & 3.44 & 2.36 & 4.58 & 6.95 & 3.81 & 4.27 & 3.38 & 7.46 & 3.10 & 4.82 & 1.99 & 0.59 & 2.05 & 2.31 & 1.66 & 3.58 \\
		{ GRNet \cite{xie2020grnet} } & 1.61 & 4.66 & 3.10 & 4.72 & 5.66 & 4.61 & 4.85 & 3.53 & 7.82 & 2.96 & 4.58 & 2.97 & 1.28 & 2.24 & 2.11 & 1.61 & 3.87 \\
		{ NSFA \cite{zhang2020detail} } & 1.51 & 4.24 & 2.75 & 4.68 & 6.04 & 4.29 & 4.84 & 3.02 & 7.93 & 3.87 & 5.99 & 2.21 & 0.78 & 1.73 & 2.04 & 2.14 & 3.77 \\
		{ SnowFlakeNet \cite{xiang2021snowflakenet} }& 1.16 & 3.32 & 2.41 & 2.97 & 5.46& 3.58 & 3.82 & 2.72 & 6.72 & 2.80 & 4.10 & 1.80 & 0.52 & 1.90 & 1.82 & 2.70 & 3.18\\
		{ VRCNet \cite{pan2021variational} } & {1.15} & 3.20 & \textbf{2.14} & 3.58 & 5.57 & 3.58  & 4.17 & 2.47 & 6.90 & 2.76 & 3.45 & 1.78 & 0.59 & 1.52 & 1.83 & 1.57 &3.06 \\ \hline
		{ \textbf{Ours}} & \textbf{0.74 }& \textbf{2.94 }& {2.25}  & \textbf{2.78} & \textbf{2.54} & \textbf{2.87}  & \textbf{2.84 }& \textbf{2.00} & \textbf{5.24} & \textbf{1.98 } & \textbf{2.87} & \textbf{1.67} & \textbf{0.45} &\textbf{1.45}&\textbf{ 1.23} & \textbf{0.92} & \textbf{2.27}\\
		\hline
	\end{tabular}
	\caption{Completion results (L2 Chamfer distance $\times 10^4$)  on  MVP  dataset (16,384 points).}
	\label{tab_MVP_16384_cd}
\end{table*}

\begin{table*}[t]
	\centering
	\setlength\tabcolsep{1.3pt}
	\small
	\begin{tabular}{l | cccccccc  cccccccc | c  }
		\hline 
		
		{Method} & {airp.} & {cabinet} & {car} & {chair} & {lamp} & {sofa} & {table} & {waterc.} & {bed} & {bench} & {books.} & {bus} & {guitar} & {motor.} & {pistol} & {skateb.}   & {Avg.}\\
		
		\hline \hline
		{ PCN} \cite{yuan2018pcn} & 		0.82 & 0.61 & 0.69 & 0.52 & 0.46 & 0.55& 0.65 & 0.63& 0.45 & 0.69   & 0.55 & 0.78 & 0.91 & 0.67 & 0.77  & 0.86 & 0.638 \\
		{ TopNet}\cite{tchapmi2019topnet} & 	0.79 & 0.62  & 0.61  & 0.44 & 0.39& 0.51& 0.64 & 0.61 & 0.41 & 0.68  & 0.52 & 0.77 & 0.87 & 0.62 & 0.73& 0.84& 0.601  \\
		{ MSN  \cite{liu2020morphing}} &0.88 & 0.69 & 0.69 & 0.60 & 0.60  & 0.63 & 0.73 & 0.80& 0.57 & 0.80 & 0.64 & 0.81& 0.94& 0.73& 0.81 & 0.89 & 0.710 \\
		{ Wang et. al. \cite{wang2020cascaded} } & 0.90 & 0.69& 0.73 & 0.67  & 0.68  & 0.64 & 0.75 & 0.74 & 0.60  & 0.80 & 0.66 & 0.80  & 0.93 & 0.77  & 0.84  & 0.90 & 0.740  \\
		{ ECG \cite{pan2020ecg}} &  	0.91 & 0.68 & 0.72& 0.68 & 0.73  & 0.65  & 0.77 & 0.75 & 0.64  & 0.82  & 0.71 & 0.80  & 0.95 & 0.78 & 0.84 & 0.90 & 0.753 \\
		{ GRNet \cite{xie2020grnet}} & 0.85  & 0.58 & 0.65 & 0.64 & 0.71 & 0.58  & 0.69  & 0.72  & 0.59 & 0.77 & 0.64 & 0.68 & 0.87 & 0.74& 0.79 & 0.85 & 0.692 \\
		{ NSFA \cite{zhang2020detail} } & 0.90  & 0.69  & 0.72 & 0.74 & 0.78 & 0.71 & 0.82 & 0.80& 0.69 & 0.85 & 0.75 & 0.82& 0.93 & 0.82& 0.86 & 0.89 & 0.783  \\
		{ SnowFlakeNet \cite{xiang2021snowflakenet} } & 0.92  & 0.70  & 0.73  & 0.72 & 0.78 & 0.70 & 0.80 & 0.79 & 0.68 & 0.85 & 0.73 & 0.82  & 0.95  & 0.80  & 0.87& 0.92 & 0.782  \\
		{ VRCNet \cite{pan2021variational} } & {0.93} & 0.72 & 0.76 & 0.74  & 0.79 & 0.70  & 0.81 & 0.80 & 0.65 & 0.86 & 0.76 & 0.83 &{ 0.96  } & 0.83 & 0.89 & { 0.93 } & 0.796  \\ \hline
		{ \textbf{Ours}} & \textbf{0.94} & \textbf{0.74} &  \textbf{0.75}& \textbf{0.77}& \textbf{0.84}& \textbf{0.74}& \textbf{0.82}&\textbf{0.82}&\textbf{0.72} & \textbf{0.87} &\textbf{0.77} &\textbf{0.84 } &{\textbf{0.97}} &\textbf{0.85} &\textbf{0.90} &{\textbf{0.93}} & \textbf{0.814}\\

		\hline
	\end{tabular}
	\caption{Point cloud completion results (F-Score@ 1\%,  higher is better) on  MVP  dataset (16,384 points).}
	\label{tab_MVP_16384_f1}
\end{table*}

\noindent	\textbf{Multi-scale Point Deconvolution (MPD) Block}.
After obtaining the modulated feature, 
we use it for point cloud refinement.
As shown in Figure \ref{prnet},
we choose snowflake point deconvolution \cite{xiang2021snowflakenet} as our base operation,
which learns the point displacement progressively to refine the shape.
For the $i$-th block,
it firstly constructs the input feature $\mathcal{Q}_{i}$,
by concatenating
the predicted point cloud $\mathcal{P}_{i}$ with its global feature from PointNet \cite{qi2017pointnet}.
Second,
it sends $\mathcal{Q}_{i}$ and $\mathcal{\hat{K}}_{i}$ into the skip transformers \cite{xiang2021snowflakenet}  for point relation learning,
where
$\mathcal{\hat{K}}_{i}$ is the modulated displacement feature from the previous block.
The output of skip transformers is denoted as $\mathcal{H}_{i}$.
Third,
it upsamples $\mathcal{H}_{i}$ as displacement feature $\mathcal{K}_{i+1}$ for next layer and generates the coordinate variation $\Delta \mathcal{P}_{i+1}$ in the current block,
by point-wise splitting operation in \cite{xiang2021snowflakenet}.
Finally, 
it adds the coordinate variation $\Delta \mathcal{P}_{i+1}$ with the duplicated previously-predicted point cloud $\mathcal{P}_{i}$,
which generates the refined point cloud $\mathcal{P}_{i+1}$ in the current block.

Note that,
our MPD block follows the refinement style of snowflake point deconvolution.
But differently,
it is equipped with our semantic-guided modulation block and multi-scale relation learning.
First,
we use the modulated feature $\mathcal{\hat{K}}_{i}$ of SGM block as input to skip transformer,
instead of using the original displacement feature $\mathcal{K}_{i}$.
In this way,
we flexibly integrate semantic category information into relation learning,
which enhances the shape-context feature $\mathcal{H}_{i}$ for discirminative refinement.
Second,  
we introduce multi-scale learning strategy for skip transformers,
instead of building point relations in a single scale.
For example, 
we use skip transformers with two different local regions
to learn shape context features
$\mathcal{H}_{i}^1$ and $\mathcal{H}_{i}^2$.
Next, 
we fuse these context features to obtain the displacement feature $\mathcal{K}_{i+1}$ in the current block,
\begin{equation}
\mathcal{K}_{i+1} =  \operatorname{MLP}( {\varphi}(\mathcal{H}_{i}^1 \oplus \mathcal{H}_{i}^2))
\end{equation}
where $\oplus$ is the concatenation operation,
and $\varphi$ is the point-wise splitting operation \cite{xiang2021snowflakenet} for upsampling.
Via such multi-scale design,
our MPD block can effectively capture complicated shape variations in the point clouds. 


\begin{table*}[t]
	\centering
	\small
	\setlength\tabcolsep{9pt}
	\begin{tabular}{l| cccccccc|c}
		\hline 
		{ Methods } &   { Plane } &  { Cabinet } &  { Car } &  { Chair } &  { Lamp } &  { Couch } &  { Table } &  { Boat }  &{ Avg. } \\
		\hline \hline
		{ FoldingNet \cite{yang2018foldingnet} }  & 9.49 & 15.80 & 12.61 & 15.55 & 16.41 & 15.97 & 13.65 & 14.99 & 14.31\\
		{ TopNet \cite{tchapmi2019topnet} } & 7.61 & 13.31 & 10.90 & 13.82 & 14.44 & 14.78 & 11.22 & 11.12 & 12.15 \\
		{ AtlasNet \cite{groueix2018papier} } & 6.37 & 11.94 & 10.10 & 12.06 & 12.37 & 12.99 & 10.33 & 10.61  & 10.85 \\
		{ PCN\cite{yuan2018pcn}}  & 5.50 & 22.70 & 10.63 & 8.70 & 11.00 & 11.34 & 11.68 & 8.59 & 9.64\\
		{ GRNet \cite{xie2020grnet} } & 6.45 & 10.37 & 9.45 & 9.41 & 7.96 & 10.51 & 8.44 & 8.04 & 8.83 \\
		{ CDN \cite{wang2020cascaded} } & 4.79 & 9.97 & 8.31 & 9.49 & 8.94 & 10.69 & 7.81 & 8.05 & 8.51  \\
		{ PMP-Net\cite{wen2021pmp}}  & 5.65 & 11.24 & 9.64 & 9.51 & 6.95 & 10.83 & 8.72 & 7.25 & 8.73\\
		{ NSFA \cite{zhang2020detail} } & 4.76 & 10.18 & 8.63 & 8.53 & 7.03 & 10.53 & 7.35 & 7.48 & 8.06 \\
		{ PoinTr \cite{yu2021pointr} } & 4.75 & 10.47 & 8.68 & 9.39 & 7.75 & 10.93 & 7.78 & 7.29  & 8.38\\
		{ SCRN \cite{wang2021cascaded}} & 4.80 & 9.94 & 9.31 & 8.78 & 8.66 & 9.38 & 7.20 & 7.91 & 8.29\\
		{ SnowFlakeNet \cite{xiang2021snowflakenet}} & \textbf{4.29} & 9.16 & 8.08 & 7.89 & \textbf{6.07} & 9.23 & 6.55 & 6.40 & 7.21 \\
		\hline
		{ \textbf{Ours}}  &4.34 & \textbf{9.02} & \textbf{7.90} &\textbf{ 7.41 }& 6.35 & \textbf{8.52} & \textbf{6.32 }&\textbf{6.26} & \textbf{7.02} \\
		\hline
	\end{tabular}
	\caption{Comparisons on PCN dataset (L1 Chamfer distance $\times 10^3$). }
	\label{tab:results_pcn}
\end{table*}

\begin{table}[h]
	\centering
	\setlength\tabcolsep{1.3pt}
	\small
	\begin{tabular}{l|cc|cc|cc|cc}
		\hline
		\multirow{2}{*}{ {  {\# Points} }} & \multicolumn{2}{|c|}{ {2,048}} & \multicolumn{2}{|c|}{ {4,096}} & \multicolumn{2}{|c|}{ {8,192}} & \multicolumn{2}{|c}{ {16,384}} \\
		\cline { 2 - 9 } &   { CD } &   { F1 } &   { CD } &   { F1 } &   { CD } &   { F1 } &   { CD } &   { F1 }\\ 
		\hline \hline
		{ PCN \cite{yuan2018pcn} } &   {9.77} &  {0.320 }&  {7.96} &  {0.458} & {6.99} &  {0.563} &  {6.02} &  {0.638} \\
		{ TopNet \cite{tchapmi2019topnet} } &  {10.11} &  {0.308} &  {8.20 }&  {0.440} &  {7.00} &  {0.533} &  {6.36} &  {0.601} \\
		{ MSN  \cite{liu2020morphing} } &  {7.90} &  {0.432 }&  {6.17 }&  {0.585} &  {5.42} &  {0.659} &  {4.90} &  {0.710} \\
		{ Wang et. al. \cite{wang2020cascaded} } &  {7.25} &  {0.434} &  {5.83} &  {0.569} &  {4.90} &  {0.680 }&  {4.30} &  {0.740} \\
		{ ECG\cite{pan2020ecg}} &  {6.64} &  {0.476 }& {5.41} &  {0.585} &  {4.18 }&  {0.690} &  {3.58} &  {0.753}\\
		{ VRCNet \cite{pan2021variational}} &  {5.96} &  {0.499} &  {4.70} &  {0.636} &  {3.64} &  {0.727} &  {3.06} &  {0.796 }\\
		{ SnowFlakeNet \cite{xiang2021snowflakenet}} &  {6.05} &  {0.500} & {4.77}  & {0.651} &  {3.80} & {0.747} &  {3.18} &   {0.782} \\
		\hline
		{ \textbf{Ours} } &  {\textbf{5.10}}  &   {\textbf{0.526}}  &  {\textbf{3.49}}   & {\textbf{0.682}}   & {\textbf{3.14}}     &  {\textbf{0.756}}  & {\textbf{2.27}}   & {\textbf{0.814}}     \\ 
		\hline
	\end{tabular}
	\caption{Comparisons on MVP dataset with various resolutions.}
	\label{tab:results_mvp}
\end{table}

\section{Experiments
	\label{sec:exp}
}

\begin{figure*}[t]
	\centering
	\includegraphics[width=0.95\linewidth]{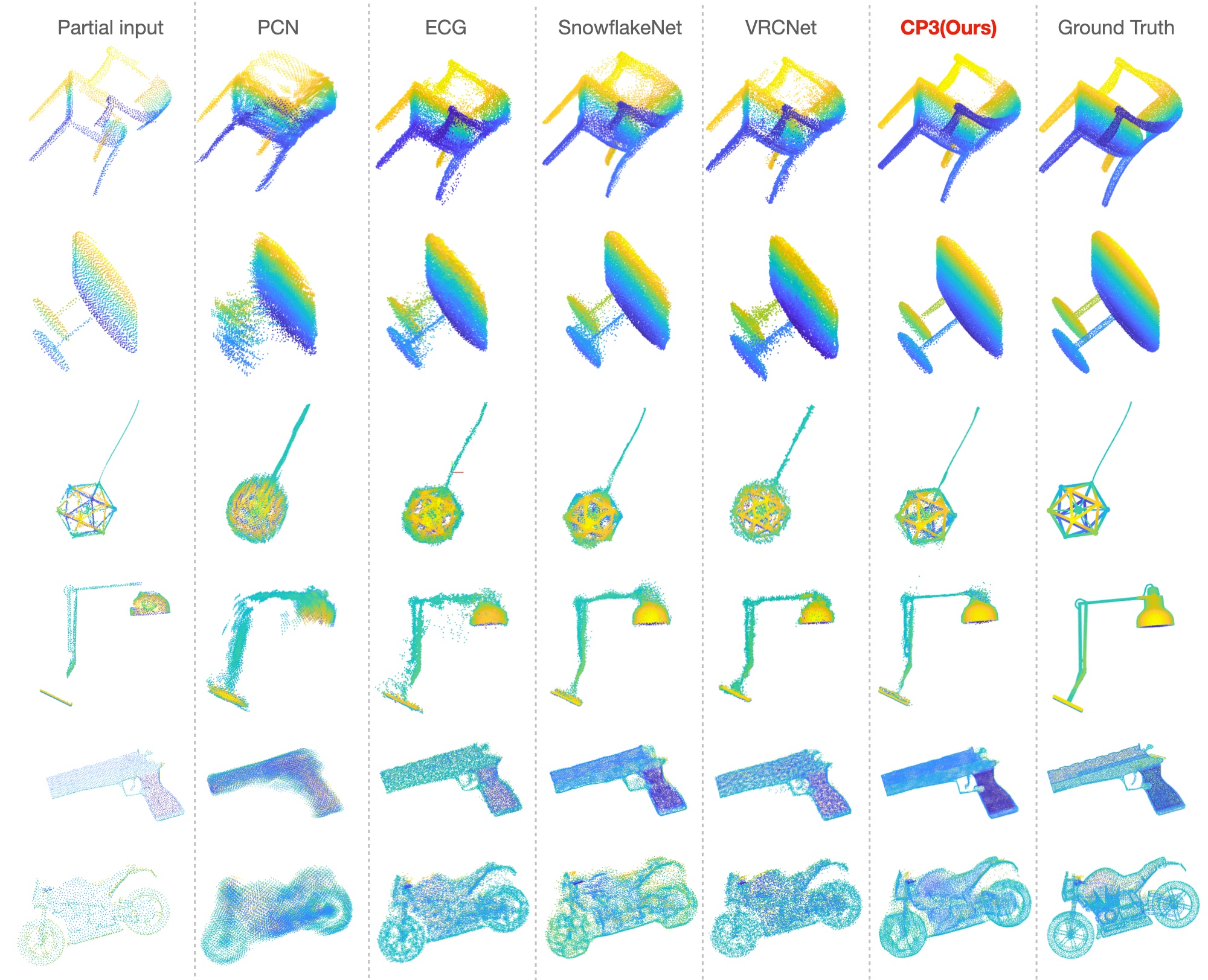}
	\caption{Qualitative completion results  on the MVP dataset by different methods. }
	\label{vis_ablation}
\end{figure*}

\subsection{Implementation Details}
In this subsection, we will introduce the specific implementation details of pretrain, prompt and predict in our CP3 paradigm, as well as the training details of the whole framework.
Firstly, we use inverse prompt function to construct the pretraining pairs, where the threshold $\Gamma$ of IOI pretraining is set to 0.9.
Then we pretrain the generation network in the self-supervised way.
After training the generation stage,
we fix it to extract the global feature and point-wise feature of the last layer,
and feed them to train the refinement stage.
In semantic-guided prediction stage, three MPD blocks are used, as illustrated in Figure \ref{prnet}, to recover the local details iteratively. 
The proposed approach also enjoys the flexibility when dealing with different resolutions of datsets, thanks to the proposed MPD block.
For example,  for 2,048 points of imperfect point cloud and 16,384 points of target point cloud,  we can set the multipliers of three MPD blocks to  $[1,1,8]$ and  choose  local regions of multi-scale transformer with 12 and 24..
We apply KL divergence and CD loss in the generation stage, witrh the weight setting in  \cite{pan2021variational}, respectively. 
In the refinement stage, only CD loss is used.


We train the method using  4 RTX 8000 GPUs, with a batch size of 64. 
The optimizer is  Adam optimizer with initial learning rate $ 1 \times 10^{-4}$ (decayed by 0.7 every 40 epochs). 
As for  resource usage, our modulation module contains only less than 43k parameters. 
The model parameters / inference time of our SCRNet is 20.7M / 0.52s, PoinTr \cite{yu2021pointr} is 30.9M / 0.65s, SnowFlakeNet \cite{xiang2021snowflakenet} is 19.2M / 0.61s, but we have the best completion result (as Table \ref{tab:results_mvp} and \ref{tab:results_pcn} shows).


\subsection{Comparison with State-Of-The-Art}

\textbf{Evaluation on MVP Dataset.}
The MVP dataset consists of multiple CAD scans, selected from ShapeNet \cite{wu20153d}.
Compared with other datasets \cite{yuan2018pcn,tchapmi2019topnet}, the MVP dataset is more challenging, due to the rich categories and incomplete variations  caused by partial point clouds rendered from 26 views.
It is worth noting that the relative poses between the 26 cameras are fixed, but the first camera pose is selected at random, which amounts to a random rotation of all 26 camera poses.
In terms of data size, MVP has 62,400 shape pairs for training and 41,600 shape pairs for testing.
In order to accurately evaluate the quality of the completions at different resolutions, it also provides complete point clouds with different resolutions, including 2048 (1x), 4096 (2x), 8192 (4x) and 16384 (8x).
For evaluation, we follow the same  metrics with VRCNet \cite{pan2021variational} to fairly compare our method with other methods,
where  L2 version of Chamfer Distance (CD)  \cite{tchapmi2019topnet} is used to evaluate average closest point distance, and  F-score \cite{knapitsch2017tanks} evaluates the  distance between object surfaces.

The evaluated CD and F-score for all evaluated methods (16,384 points) are reported in Table \ref{tab_MVP_16384_cd} and \ref{tab_MVP_16384_f1}, respectively.
Additionally,
we re-implement the official code of SnowFlakeNet \cite{xiang2021snowflakenet} for a fair comparison.
We can find that our method can achieve the best performance in CD loss with 2.27 and F-score @ 1\% with 0.814 for 16384 (8x) points.
Our method delivers significant performance improvements  on the  categories  that have close geometric characteristics (e.g., table, chair and sofa).
Moreover,
we also compare our methods with existing methods that support multi-resolution completion in Table \ref{tab:results_mvp}. 
Our method outperforms all the other methods with a large margin.

\quad

\noindent	\textbf{Evaluation on PCN Dataset.}
Another widely-used dataset for point cloud completion is PCN dataset \cite{yuan2018pcn}, which is a subset of ShapeNet dataset  \cite{wu20153d}  and contains 8 categories.
The incomplete shapes are generated by backprojecting complete shapes into 8 different partial views. 
For each complete shape, 16,384 points are evenly sampled from the shape surface.
To  verify the effectiveness of our method and fairly compare our our method with other methods,
we conduct experiments on this dataset by following the experimental setting as used in \cite{yuan2018pcn,yu2021pointr,xiang2021snowflakenet},
where 
we adopt the L1 version of chamfer distance as the evaluation metric, which follows the same practice as previous methods \cite{yuan2018pcn,yu2021pointr,xiang2021snowflakenet}.
The results are summarized in Table \ref{tab:results_pcn}, our method achieves the best performance with a chamfer distance of 7.02 and outperforms other state-of-the-art methods in most categories. 
Thanks to the CP3 training paradigm, our approach lowering the CD by  $0.19 \times10^3$ compared with SnowFlakeNet  \cite{xiang2021snowflakenet}, demonstrating strong robustness when handling  cases of easily confusing categories.


\quad

\noindent\textbf{ Qualitative Analysis of different methods. }
To make comparisons under the challenging situations,
we visualize the completion results of different methods on MVP dataset with 16,384 points.
The qualitative results are shown in Figure \ref{vis_ablation}.
Compared to other methods, 
our proposed CP3 can recover better complete shapes with fine details.
Our method largely improves fine-graind completion, especially for classes that have close geometric clues (e.g. table and chair).
Our method also obtains satisfactory performance on objects with challenging partial areas, for example, the lamp shown in Figure  \ref{vis_ablation}  can be recovered with fine details.
More examples can be observed in Figure  \ref{vis_ablation} . Our method shows a uniform local density with much less noise (see trigger in pistol and tires in motorbike). 

\subsection{Analysis
	\label{sec:ablation} 
}

\begin{figure}[!t]
	\centering
	\includegraphics[width=1\linewidth]{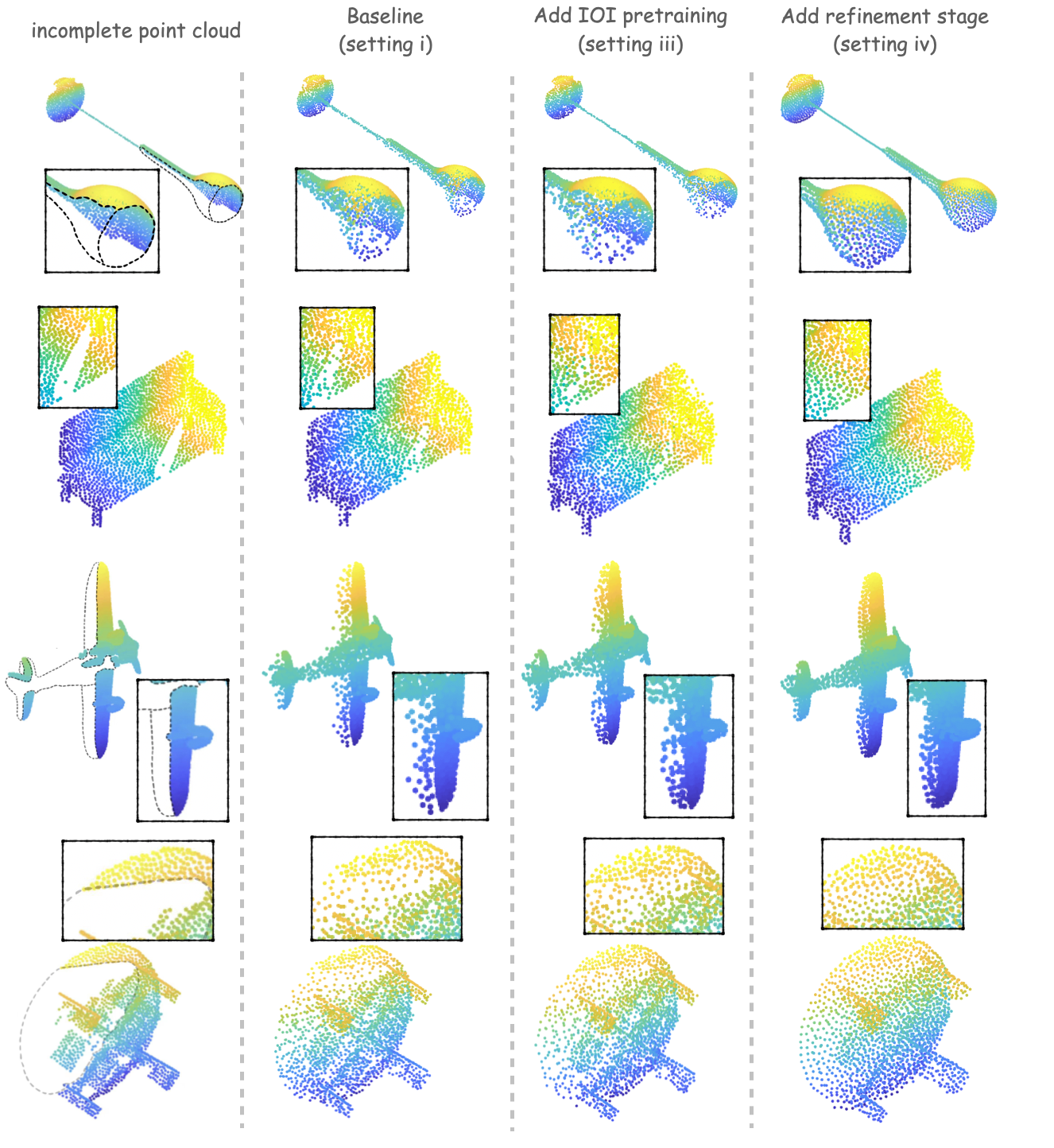}
	\caption{Qualitative completion results   by different ablation settings. }
	\label{fig:sup_ablation_vis}
\end{figure}

\subsubsection{Ablation Studies of Overall Framework}

\noindent\textbf{Why does  CP3 paradigm work for point cloud completion?}
In order to examine the effectiveness of our designs,
we conduct comprehensive ablation studies based our framework on MVP dataset (2048 points).
The results are summarized in Table \ref{tab:ablation_framework}.
 ($\romannumeral 1$) only has the generation stage (prompt), where the result is 5.96. 
Specifically, VRCNet \cite{pan2021variational} is used as the generation network.
Based on the prompt,  ($\romannumeral 2$) adds the  predict stage to point cloud refinement.
The  result improves to 5.80 due to our semantic modulation strategy for discriminative shape recovery.
($\romannumeral 3$) and ($\romannumeral 4$) add the IOI pretraining paradigm.
When the completion model only includes the generation stage, the result is improved to 5.69.
While we use the entire framework of CP3, the result can be improved to 5.10.
It can be concluded that our proposed CP3 training paradigm can greatly improve the performance of point cloud completion.

{Can IOI pretraining be effective on refinement  or semantic guided modulation on generation?}
 To this end, we use the semantic guided modulation block to modulate the local representation in generation (prompt stage).
  The results are listed in Table \ref{tab:ablation_framework2}($\romannumeral 5$), which drops to 5.93 from 5.69, 
  indicating that it is not appropriate to apply semantic guided modulation on generation for pretraining and prompting,
  by the reason of that  generation is concerned with the mapping of partial to complete, and there is less  semantic distinction among the incomplete point clouds.
  Based on setting ($\romannumeral 4$),we also apply the IOI pretraining to the semantic refinement network, the result drops to  5.30 (setting  $\romannumeral 6$), which  shows that IOI pretraining has no contribution to the refinement  (predict stage).
  Considering the key of refinement is to recover fine-grained shape details with the imperfect completion,
  whereas IOI pretraining trackles the problem of the generation stage.
 These two results indicate that the generation and refinement have different fundamental problems about how to use point cloud  effectively,  and  our proposed IOI pretraining and semantic guided modulation  are designed for different stages respectively.

\quad

\noindent\textbf{Qualitative visualizations of framework ablations.}
To qualitatively validate our framework, we conduct a visual comparison of some settings  in Table \ref{tab:ablation_framework},
the visualizations are shown in Figure \ref{fig:sup_ablation_vis}.
We can clearly observe that completions of  setting $\romannumeral 1$   have some defects, e.g., the  density of incomplete areas and complete areas is inhomogeneous, noisy points are generated in the hollowed out areas and some partial regions can  not be completely filled in.
With the IOI pretraining $\romannumeral 3$ , the large missing areas can  be completed more accurately.
e.g., the incomplete areas of the bench and lamp are more complete in setting  $\romannumeral 3$.
When adding our semantic conditional network for refinement (setting  $\romannumeral 4$), the density problem can be solved effectively and the missing areas can be more complete.

\begin{table}[t]
	\small
	\setlength\tabcolsep{11pt}
	\centering
	\begin{tabular}{l|ccc|c}
		\hline 
		CP3 &  Pretrain (IOI) & Prompt  & Predict & CD\\
		\hline $\romannumeral 1$  & $\times$ & \checkmark & $\times$ & 5.96 \\
		$\romannumeral 2$ & $\times$  &  \checkmark  &  \checkmark  & 5.80 \\
		$\romannumeral 3$ & \checkmark  &  \checkmark & $\times$ & 5.69 \\
		$\romannumeral 4$ &   \checkmark  &  \checkmark  & \checkmark  &  5.10 \\
		\hline
		
	\end{tabular}

	\caption{Ablation studies of  the CP3 framework  on MVP dataset with 2048 points (CD  multiplied by $10^4$),  ``$\checkmark$" means that we use the corresponding strategies, while ``$\times$" means  that the relevant strategies is not applicable.}
	\label{tab:ablation_framework}
\end{table}

\begin{table}[t]
	\small
	\setlength\tabcolsep{13pt}
	\centering
	\begin{tabular}{l|ll|c}
		\hline 
		CP3 & generation  & refinement & CD\\\hline
		$\romannumeral 2$ &  \checkmark & $\times$ & 5.69 \\
		$\romannumeral 5$ &  \checkmark +SGM & $\times$ & 5.93 \\
		$\romannumeral 4$ &  \checkmark  & \checkmark  &  5.10 \\
		$\romannumeral 6$ &  \checkmark  & \checkmark+IOI pretrain    &  5.30 \\
		\hline
		
	\end{tabular}
	\caption{Ablation studies of  the CP3 framework  on MVP dataset with 2048 points (CD  multiplied by $10^4$). ``SGM" means semantic guided modulation.}
	\label{tab:ablation_framework2}
\end{table}

\begin{table}[t]
	\small
	
	\setlength\tabcolsep{3pt}
	\centering
	\begin{tabular}{l | l  l| l}
		\hline
		& Pretraining input   & Pretraining Target  & CD    \\ \hline 
		Pre-1 & Mirrored Incompletion  & Completion & 7.59\\
		Pre-2  & Jittered Incompletion & Completion & 7.16 \\
		Pre-3 & Incompletion & Completion &  5.96\\ 
		Pre-4 &  Incompletion & Incompletion & 5.78 \\
		Pre-5 & Incompletion-of-Incompletion & Completion & 6.03 \\ 
		Pre-6 & {hybrid training as SCRN\cite{wang2021cascaded}} &  & 5.75\\
		Pre-7 & Incompletion-of-Incompletion  & Incompletion & 5.69\\
		\hline
	\end{tabular}
	\caption{Ablation studies of IOI pretraining: different pretraining settings.}
	
	\label{tab:ablation_generation}
\end{table}

\begin{table}[t]
	\centering
	\setlength\tabcolsep{10 pt}
	\small
	\begin{tabular}{ l |c c }
		\hline 
		Generation Network  & w/o IOI Pre. & w IOI Pre. \\ \hline
		PCN \cite{yuan2018pcn} & 9.77 & 8.51  \\ 
		ECG \cite{pan2020ecg} & 6.64 &   6.34 \\
		VRCNet \cite{pan2021variational} & 5.96 & 5.69 \\
		\hline
	\end{tabular}
	\caption{Ablation studies of IOI pretraining: different generation networks.}
	\label{tab:ablation_IOI_other_methods}
\end{table}

\begin{table}[!t]
	\centering
	\renewcommand\arraystretch{0.7}
	\setlength\tabcolsep{15 pt}
	\small
	\begin{tabular}{ l | c}
		\hline 
		 & CD \\
		\hline
	baseline & 5.96\\
		IOI augmentation & 5.82 \\
		IOI pretraining & 5.69\\
		\hline
	\end{tabular}
	
	\caption{Comparison of IOI pretraining and augmentation.}
	\label{tab:ablation_pretraining_augmentation}
\end{table}

\begin{table}[t]
	\centering
	\renewcommand\arraystretch{0.7}
	\setlength\tabcolsep{19 pt}
	\small
	\begin{tabular}{ l | c}
		\hline 
		Crop procedure of IOI pretraining & CD \\
		\hline
		baseline & 5.96\\
		projection based crop & 5.69 \\
		multiple local patches crop & 5.70\\
		\hline
	\end{tabular}
	
	\caption{Ablation studies of IOI pretraining:  different crop procedure  of IOI sampling.}
	\label{tab:ablation_crop_procedure}
\end{table}

\begin{table}[t]
		\centering
		\renewcommand\arraystretch{0.7}
		\setlength\tabcolsep{9 pt}
		\small
		\begin{tabular}{ c | c cccc}
			\hline 
			 	threshold:  $\Gamma$  &	0.1 & 0.3 & 0.5 & 0.7 & 0.9 \\
			 		\hline
			 		CD &  5.80& 5.78 & 5.73 & 5.71 &  5.69 \\
			\hline
		\end{tabular}

	\caption{Ablation studies of IOI pretraining:  different maximun threshold of the crop rate in projection based crop procedure.}
		\label{tab:ablation_IOI_crop}
\end{table}

\begin{table}[!t]
	\centering
	\renewcommand\arraystretch{0.7}
	\setlength\tabcolsep{5 pt}
	\small
	\begin{tabular}{ c | c c cccc}
		\hline 
		\# crop patches: $\textit{m}$  & 0&	1 & 2 & 3 & 4 & 5 \\
		\hline
		CD & 5.96& 5.78& 5.72 & 5.77 & 5.70 &  5.78 \\
		\hline
	\end{tabular}
	
	\caption{Ablation studies of IOI pretraining:  different patches of the multiple local patches crop procedure.}
	\label{tab:ablation_IOI_multi_patches}
\end{table}

\begin{figure*}[t]
	\centering
	\includegraphics[width=1\linewidth]{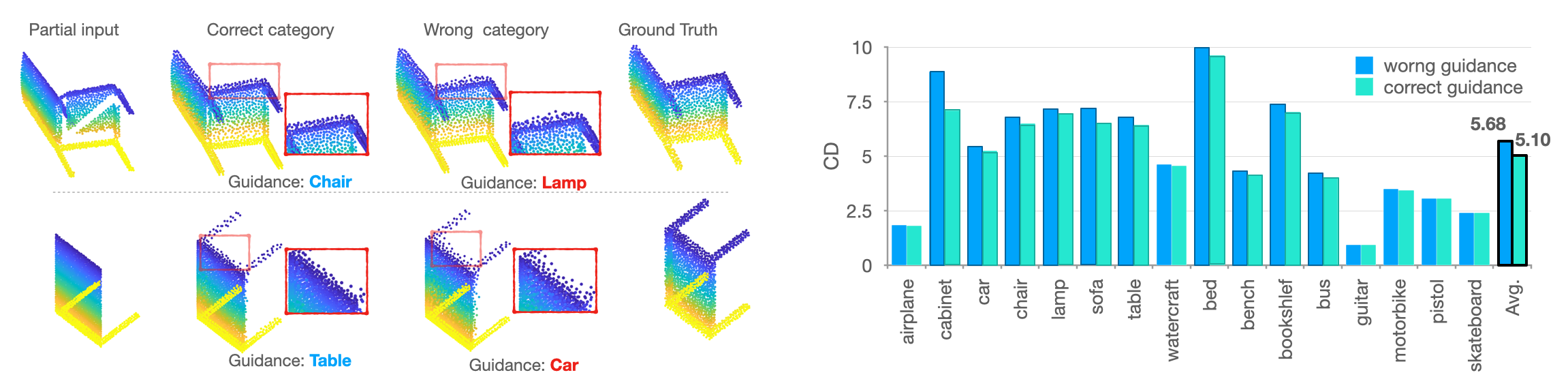}
	\caption{ Evaluating with wrong category guidance.
		Left: Qualitative evaluation results of different setting to the category guidance.
		Right:
		Comparisons of  correct/wrong  category guidance for evaluation   on  MVP  dataset  (\textit{Lower is better}).
	}
	\label{tab_MVP_test_label}
\end{figure*}

\subsubsection{Ablation Studies of IOI Pretraining}

\noindent	\textbf{Why does the IOI pretraining work?}
To verify the effectiveness of our IOI pretraining, we compare a variety of pretraining settings in Table \ref{tab:ablation_generation}.
When we use the incomplete point cloud to reconstruct itself (``Pre-4") as the pretext task, the result can improve to 5.78, because the model learns the distribution of each incomplete point cloud independently, and then plays a better  generalization in the generation stage.
Next we try to use completion as a pretext task (``Pre-5 " and ``Pre-7").
Compared with incomplete point cloud,
the incomplete of incomplete point cloud has a greater degree and more diverse partial areas, 
because it has undergone the twice incomplete treatment.
It is difficult for the network to complete the  IOI input to the complete target  (``Pre-5 ").
As a result, it drops to 6.03 from 5.96.
Therefore, it is more appropriate to process the  augment input to the original incomplete point cloud, where ``Pre-7"  improves to 5.69.
We also consider some commonly used data diversified operations in point cloud analysis, e.g., mirroring and jittering the point cloud,   demonstrating by the  ``Pre-1" and ``Pre-2".
Their results are reduced to 7.59 and 7.16 respectively, which indicates that such operations destroy the  structures of the point cloud and make the completing  excessively difficult.
As for comparison with other pretraining manners  for generation stage, we  use the same pretraining strategy as 
SCRN \cite{wang2021cascaded}.
SCRN is a hybrid training of shape completion (``IOI" to ``I") and partial reconstruction (``IOI" to ``IOI") in a \textit{parallel} manner.
To show our effectiveness,
we conduct such parallel style of SCRN as ``Pre-6 ", which gets the performance  of 5.75.
Considering such parallel pretraining may mislead the generation model,
by using the similar input but with different supervision.
In contrast,
our CP3 follows a \textit{serial} training manner from NLP, 
with ``IOI" to ``I" for pretraining and then ``I" to ``C" for prompting,
and  achieves better performance of 5.69.


%
To further explore  how IOI sampling can be used more effectively,
we  apply IOI sampling for pretraining and data augmentation of the generation network respectively.
It is worth noting that IOI augmentation involves the IOI training pairs as augmented data in the training process of the generation network.
As Table \ref{tab:ablation_pretraining_augmentation} indicates, 
without   pretraining and augmentation, the  generation network  achieves the result of 5.96 (baseline).
Using  IOI augmentation strategy, the result can be improved to 5.82,
while   IOI pretraining can achieve to 5.69.
Although IOI augmentation is effective, it does not provide as large a performance gain as  IOI pretraining.
This is due to the fact that IOI augmentation mixes different partial to complete pairs together for training, which may confuse the model.
On the other hand, 
IOI pretraining naturally circumvents this problem by  pretraining the IOI pairs and then finetuning on the original  pairs.

\quad

\noindent\textbf{Ablation studies of crop procedure in IOI pretraining. }
The IOI sampling  is not a fixed form of crop procedure, 
we offer different forms of crop methods, i.e., projection based crop and multiple local patches crop,
both can improve the performance (Table \ref{tab:ablation_crop_procedure}). 
As  for projection based crop  procedure,
we  compare the maximum threshold $\Gamma$ of the crop rate based on the IOI  pretraining, which is shown in Table \ref{tab:ablation_IOI_crop}. 
Through the series of experiments, we finally determine the  threshold  $\Gamma$  is 0.9,
and the other settings of $\Gamma$   are also valid.
Towards   the multiple local patches crop  procedure, 
we compare different numbers of local patches on IOI pretraining in Table \ref{tab:ablation_IOI_multi_patches}.
Compared to the baseline (\textit{m}=0), all setttings have different degrees of improvement, with the best results for \textit{m}=4.
From the above ablations, it can be concluded that most  reasonable sampling strategies can make IOI pretraining effective.

\quad

\noindent\textbf{Can  IOI pretraining paradigm work for other  networks? }
Here we compare the effectiveness of IOI pretraining for different  networks (e.g., PCN \cite{yuan2018pcn}, ECG \cite{pan2020ecg} and VRCNet \cite{pan2021variational}) on Table \ref{tab:ablation_IOI_other_methods}.
With IOI pretraining, 
the performance of such  mainstream completion networks  is greatly improved in varying degrees.
It can be concluded that the IOI pretraining is a general and effective pretraining paradigm for point cloud completion task.

\subsubsection{Ablation Studies of  Semantic Refinement}

\noindent\textbf{Why does the semantic refinement work?}
To validate the designs of our predict (semantic-guided refinement) stage, we first conduct three different  variations as Table \ref{tab:ablation_SCRNet} shows.
When we use the  basic refinement network without semantic-guided modulation and multi-scale transformer strategy, the result is 5.60.
With the semantic modulation, the charmer distance improves to 5.17.  
Subsequently, when adding the multi-scale strategy in the MPD blocks, the final result can achieve to 5.10.
Moreover, we investigate different  conditional information in Table \ref{tab_ablation_modulation}.
It can be concluded from the experiments that under the modulation of category semantics and global learnable semantic, the performance can be  greatly improved.
Specially, the category semantics is more effective as the conditional information.
In order to further  validate Semantic-Guided Modulation (SGM) block, we 
conduct a series of comparative experiments of whether to use the SGM block in each layer of our SCRNet, which is shown in Table \ref{tab:ablation_whether_mvp}.
It is relatively more efficient to use semantic-guided modulation blocks at the front layers.
Moreover,
for the multi-scale strategy, we conduct different settings for the transformer scale in Table \ref{tab_ablation_multi_scale}, which indicates that it is more effective to choose a larger local pattern of fusion.

\quad

\begin{figure}[!t]
	\centering
	\includegraphics[width=0.9\linewidth]{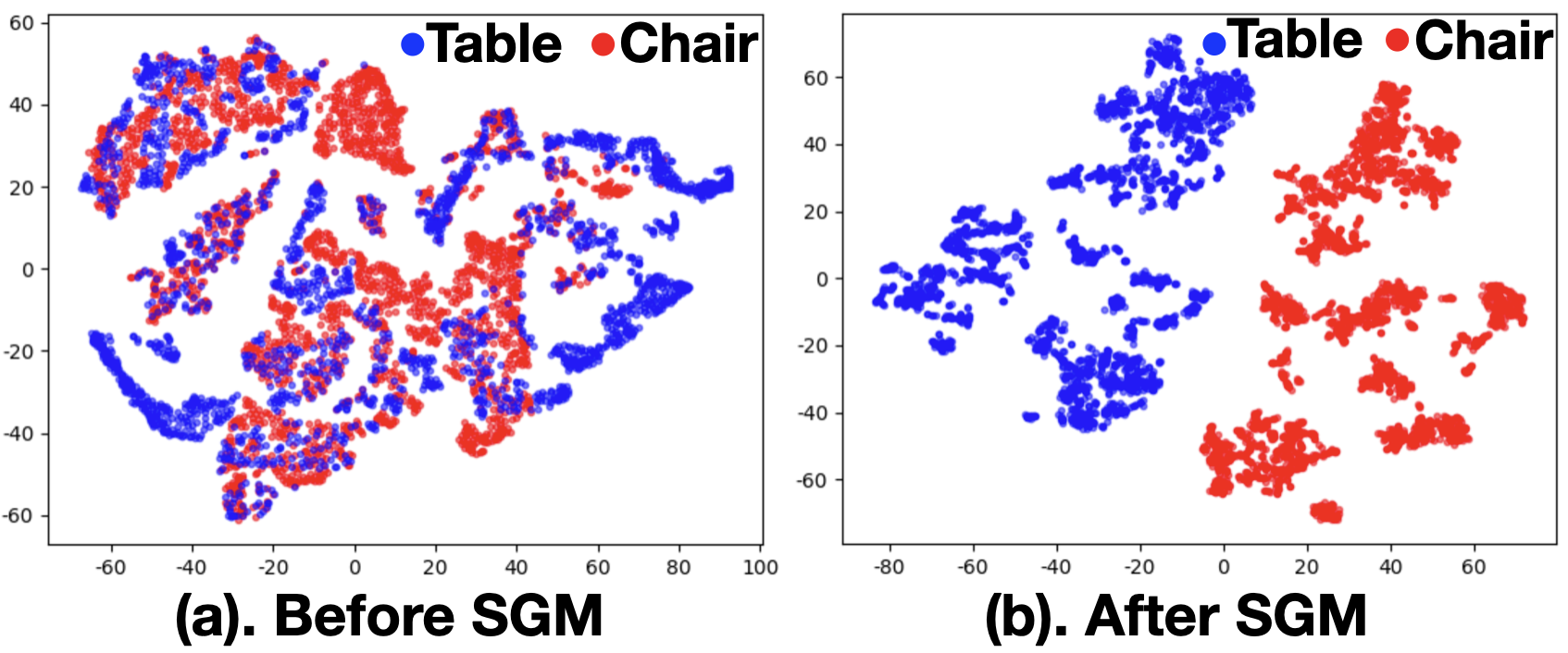}
	\caption{TSNE visualization of features  before and after semantic guided modulation. }
	\label{fig:tsne}
\end{figure}

\noindent\textbf{Why semantic label contributes to completion task?}
Analysis from the perspective of point-wise representation,
our SGM   modulates each point-wise feature with semantic label. 
This can effectively distinguish point clouds which have similar local details but belong to different categories.
In Figure \ref{fig:tsne},
we show the original point-wise feature distribution of two point clouds (chair and table)  which are  largely mixed up.
Clearly,
the  features of  two point clouds  (chair and table) are distinguished after using our SGM. 
As a result,  using semantic information to guide  fine-grained completion  becomes  easier.



\begin{table}[!t]
	\centering
	\setlength\tabcolsep{10pt}
	\small
	\begin{tabular}{ c c | c}
		\hline 
		{Semantic modulation}  & {Multiscale operation} &CD\\
		\hline
		$\times$ &  $\times$ & 5.60  \\
		$\checkmark$ &  $\times$ &  5.17 \\
		$\checkmark$ &  $\checkmark$ & 5.10 \\
		\hline
	\end{tabular}
	\caption{Ablation studies of our SCRNet for refnement:  SGM and MPD blocks in the refinement. }
	\label{tab:ablation_SCRNet}
\end{table}

\begin{table}[!t]
	\small
	\setlength\tabcolsep{13 pt}
	\centering
	\begin{tabular}{c c c | c}
		\hline
		1-st layer & 2-nd  layer  & 3-rd layer  & CD    \\ \hline 
		$\checkmark$ &$\times$ &  $\times$ &  5.17 \\ 
		$\times$  &$\checkmark$&  $\times$ &  5.18 \\ 
		$\times$ &  $\times$ & 	$\checkmark$ &  5.20 \\ 
		\hline
	\end{tabular}
	\caption{Ablation studies of our SCRNet for refnement: whether to use the semantic-guided modulation block.}
	\label{tab:ablation_whether_mvp}
\end{table}

\begin{table}[!t]
	\centering
	\setlength\tabcolsep{15 pt}
	\small
	
	\begin{tabular}{ l  l |c  }
		\hline 
		Strategy &local regions ($k$)  &  CD \\\hline
		single-scale &[12] & 5.17\\ 
		multi-scale & [12, 24] & 5.10 \\ 
		multi-scale  &[6,12] & 5.23 \\ 
		\hline
	\end{tabular}
	
	\caption{Ablation studies of our SCRNet for refnement:  $k$-nearest neighbours  of the multi-scale strategy. }
	\label{tab_ablation_multi_scale}
\end{table}

\quad

\noindent	\textbf{Evaluating with wrong category guidance.}
To verify the effectiveness of semantic-guided modulation on different categories,
we  use  model $\romannumeral 4$  in Table \ref{tab:ablation_framework} to evaluate our CP3 with different strategies of category guidance.
The results are shown in Figure \ref{tab_MVP_test_label}.
When we randomly replace the categories of all test point clouds, and the CD loss becomes larger (i.e., 5.68). 
For further analysis, we show the test results for all categories.
We can see that,
for a small number of categories with high discrimination (e.g., airplane, watercraft, guitar), 
their completions are not greatly affected by the wrong guidance.
While for most confusing categories with similar structures, (e.g.,  chair, sofa and table), 
their evaluation results are heavily deteriorated by the wrong guidance.
Moreover,
The visualization of different guidance in Figure \ref{tab_MVP_test_label} can also verify this viewpoint.
The local area  have more noisy points under the wrong guidance than the correct category guidance.
It shows that our semantic conditional modulation network is effective to reduce shape confusion in the  fine-grained completion.




\begin{figure}[!t]
	\centering
	\includegraphics[width=0.75\linewidth]{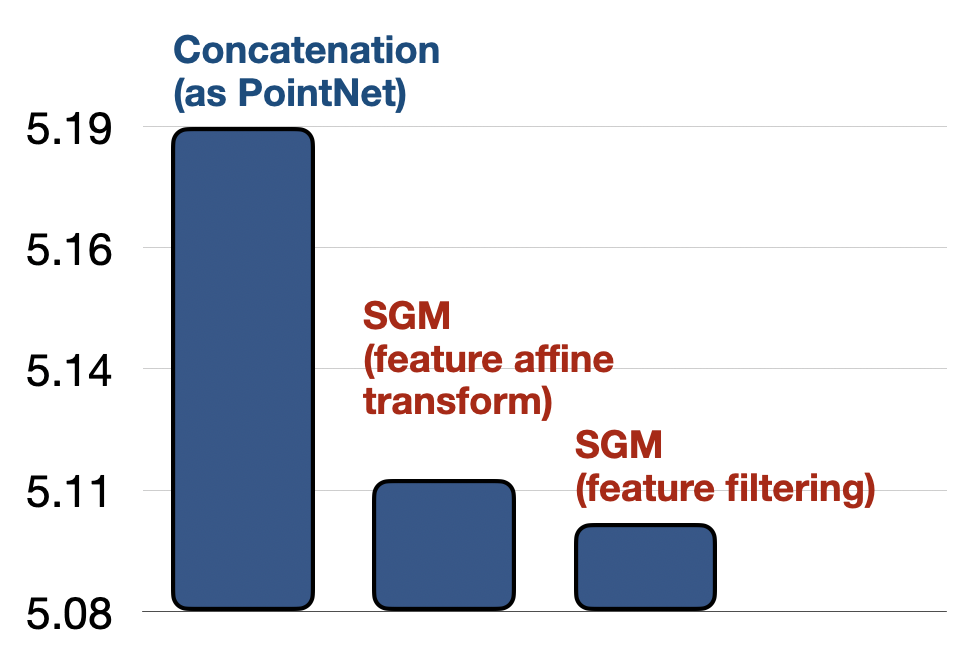}
	\caption{Comparisons of different feature modulation operations. }
	\label{fig:sup_fam}
\end{figure}

\begin{table}[!t]
	\centering
	\setlength\tabcolsep{15 pt}
	\small
	\begin{tabular}{ l |c  }
		\hline 
		Conditional information  &  CD \\\hline
		{without semantics }   & 5.60\\ 
		{category semantics } &  5.17 \\
		{global learnable semantics } & 5.25   \\
		{category  + global learnable semantics  } &  5.22\\ 
		\hline
	\end{tabular}
	\caption{Ablation studies of our SCRNet for refnement: conditional information in our semantic-guided modulation. 
	}
	\label{tab_ablation_modulation}
\end{table}

\begin{table}[!t]
	\centering
	\setlength\tabcolsep{25 pt}
	\small
	\begin{tabular}{ l | c}
		\hline 
		{Method}   &CD\\
		\hline
		PCN \cite{yuan2018pcn} & 16.56 \\
		ECG \cite{pan2020ecg} & 15.23 \\
		PoinTr \cite{yu2021pointr} & 10.31 \\
		VRCNet \cite{pan2021variational} & 9.08 \\ 
		SnowFlakeNet \cite{xiang2021snowflakenet} & 8.25\\
		\hline
		\textbf{Ours} & 7.64 \\
		\hline
	\end{tabular}
	\caption{Generalization to novel unseen categories on MVP dataset. }
	\label{tab_unseen}
\end{table}

\quad

\noindent	\textbf{Ablations of feature modulation operations.}
Based on the setting    $\romannumeral 4$ of the Table \ref{tab:ablation_framework},
we investigate different  operations for  feature modulation which are reported  in Figure \ref{fig:sup_fam}.
From the looks of it, 
when we simply concatenate categories and point features (similar as PointNet \cite{qi2017pointnet}),  
the performance is 5.19,
as a result of  ignoring the discriminative interaction between high level categories and point-wise features.
However, both of our semantic guided modulations can  provide the  performance boost,
where feature filtering achieves 5.10, considering it converts category into a dynamical filter to modulate all the point-wise features with semantic information.
Feature affine modulation gets the result of 5.11, owing to scale and shift the intermediate features  by two conditional vectors, which  affect the cluster centers of the local representation and  fine-tune the variance in the feature space.
Therefore, the model is not easy to be confused with the similar local structures of different semantic information.

\begin{figure}[t]
	\centering
	\includegraphics[width=0.95\linewidth]{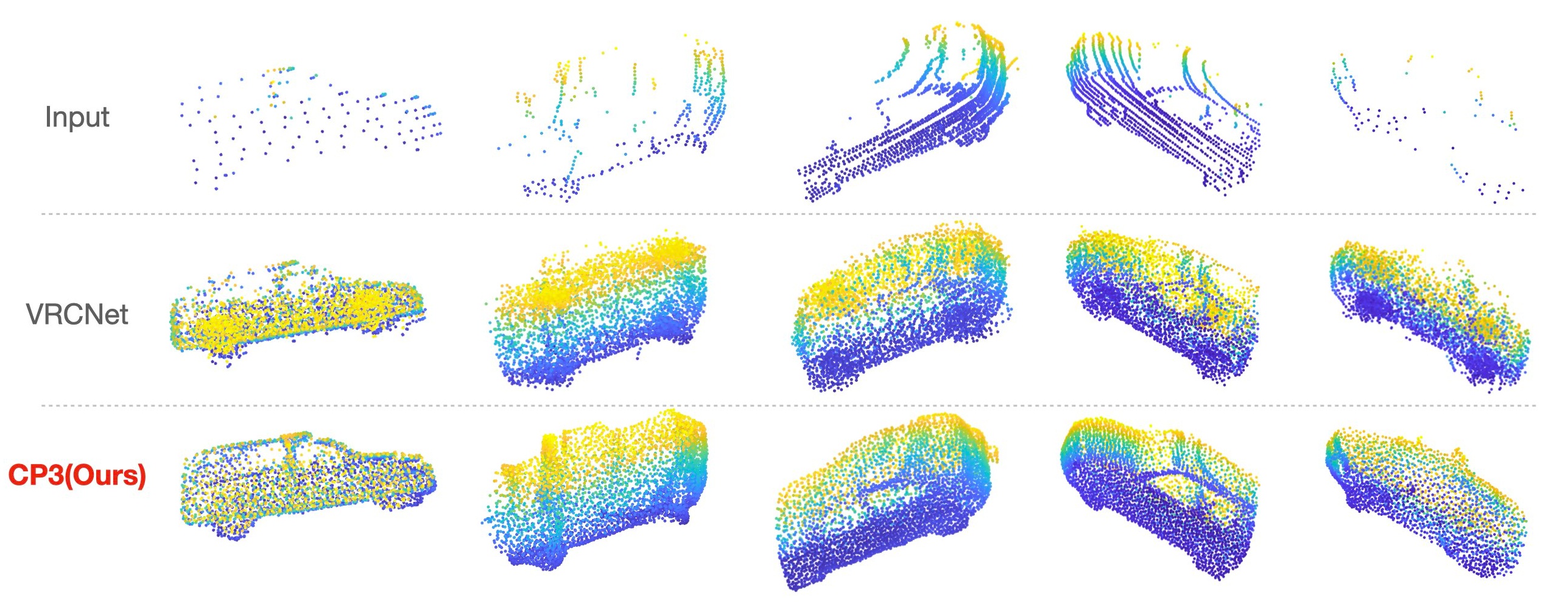}
	\caption{ Qualitative results on the KITTI dataset. Our method generates  more complete and locally detailed shapes for the real scanned cars.}
	\label{fig:sup_kitti_vis}
\end{figure}

\begin{table}[t]
	\setlength\tabcolsep{0.7pt}
	\centering
	\scriptsize
	\begin{tabular}{ l | cccccc }
		\hline 
		& PCN \cite{yuan2018pcn}  &  FoldingNet \cite{yang2018foldingnet} & NSFA\cite{zhang2020detail} &  TopNet\cite{tchapmi2019topnet}&GRNet\cite{xie2020grnet} & Ours\\
		\hline
		Consistency $\downarrow$ & 1.56 & 1.05 &  - & 0.57 & 0.31 & \textbf{0.29}\\
		Fidelity error $\downarrow$ & 0.034 & 0.036 & 0.026 & 0.031 & -  & \textbf{0.013 } \\
		\hline
	\end{tabular}
	\caption{Quantitative comparisons on KITTI dataset. }
	\label{tab:A5}
\end{table}

\subsubsection{Generalization Analysis to novel unseen categories}
To verify the generalization ability to novel unseen categories,
following  the setting of \cite{yu2021pointr},
we take the first 8 categories of MVP \cite{pan2021variational} for training and the rest 8 categories of the test dataset for evaluation.  
As shown in Table \ref{tab_unseen}, 
we compare our CP3 with some sota methods, e.g., PoinTr \cite{yu2021pointr}, VRCNet \cite{pan2021variational}, SnowFlakeNet \cite{xiang2021snowflakenet}.
Under such challenging situation, our method can achieve the best performance of 7.64.
While VRCNet only achieves 9.08 and SnowFlakeNet achieves 8.25.
It can be seen that our CP3 paradigm is also very effective for unseen categories.
This is due to the fact that IOI pretraining improves the robustness to tackle various incompletions, which is   more friendly to completing the unseen categories.

\subsubsection{Completion on real-world Kitti dataset}

In order to validate the performance of our CP3 in real-world scenarios, 
following \cite{pan2021variational,yuan2018pcn},
we fine tune our model on ShapeNet-car dataset and test on KITTI-cars \cite{geiger2012we}.
To show the performance of our CP3 in real-world KITTI-cars dataset,
we report the Fidelity error and Consistency in Table \ref{tab:A5}.
Fidelity  is  calculated from the average distance from each point in the partial input to its nearest neighbour in the output. This measures how well the input is preserved. 
Consistency is calculated from the average CD distance between the outputs of the same object in successive frames.
More details on the evaluation metrics can be found in \cite{yuan2018pcn,xie2020grnet}.
Compared to other methods, our CP3 has better consistency between the two consecutive frames and better fidelity error.
Moreover, we also show the qualitative visualization results  in Figure \ref{fig:sup_kitti_vis}, 
compared to the baseline method, our results are more complete overall and less noisy locally,
where the bodywork of the car has also been refined in detail.


\section{Conclusion}
In this paper, we creatively propose a new paradigm called CP3 for point cloud completion, including IOI pretraining, prompting (generation) and predicting (refinement).
IOI pretraining boosts the  robustness of generation network by learning a pretext task internally in a self-supervised  way.
Moreover, 
we design a novel semantic  conditional modulation for point cloud refinement,
which uses semantic information as guidance to adaptively modulate point cloud representation for discriminative recovery.
Extensive experimental results and visualizations validate the effectiveness of our approach.

\quad

\noindent	\textbf{Broader Impact}.
The core of our CP3 is a generic training paradigm,
inspired by the well-established natural language processing.
Hence,
it would be interesting to extend this paradigm for other 3D tasks like point cloud scene understanding.
Furthermore,
it is worth investigating whether it is an effective method for traditional vision tasks like image inpainting.
Hopefully,
such paradigm can leverage benefits from both CV and NLP for better representation learning in a unified manner.
\ifCLASSOPTIONcaptionsoff
  \newpage
\fi



\bibliographystyle{IEEEtran}
\bibliography{IEEEabrv,./egbib}
\begin{IEEEbiography}[{\includegraphics[width=1in,height=1.25in,clip,keepaspectratio]{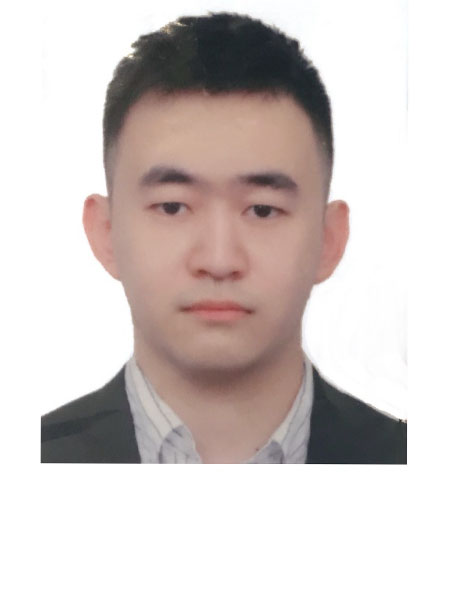}}]{Mingye Xu}
	is now a PhD student at University of Chinese Academy of Sciences, Shenzhen Institutes of Advanced Technology, Chinese Academy of Sciences, majoring in pattern recognition and intelligent system. His research interests include 3D computer vision and deep learning.
\end{IEEEbiography}
\vspace*{1em}

\vspace*{1em}
\begin{IEEEbiography}[{\includegraphics[width=1in,height=1.25in,clip,keepaspectratio]{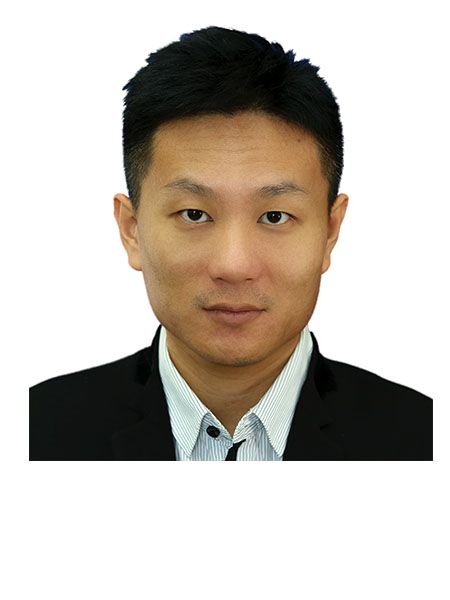}}]{Yali Wang}
	 received the Ph.D. degree in computer science from Laval University, Quebec, QC, Canada,in 2014. He is currently an Associate Professor with the Shenzhen Institutes of Advanced Technology, Chinese Academy of Sciences. His research interests are deep learning and computer vision, machine learning, and pattern recognition.
\end{IEEEbiography}
\vspace*{1em}
\begin{IEEEbiography}[{\includegraphics[width=1in,height=1.25in,clip,keepaspectratio]{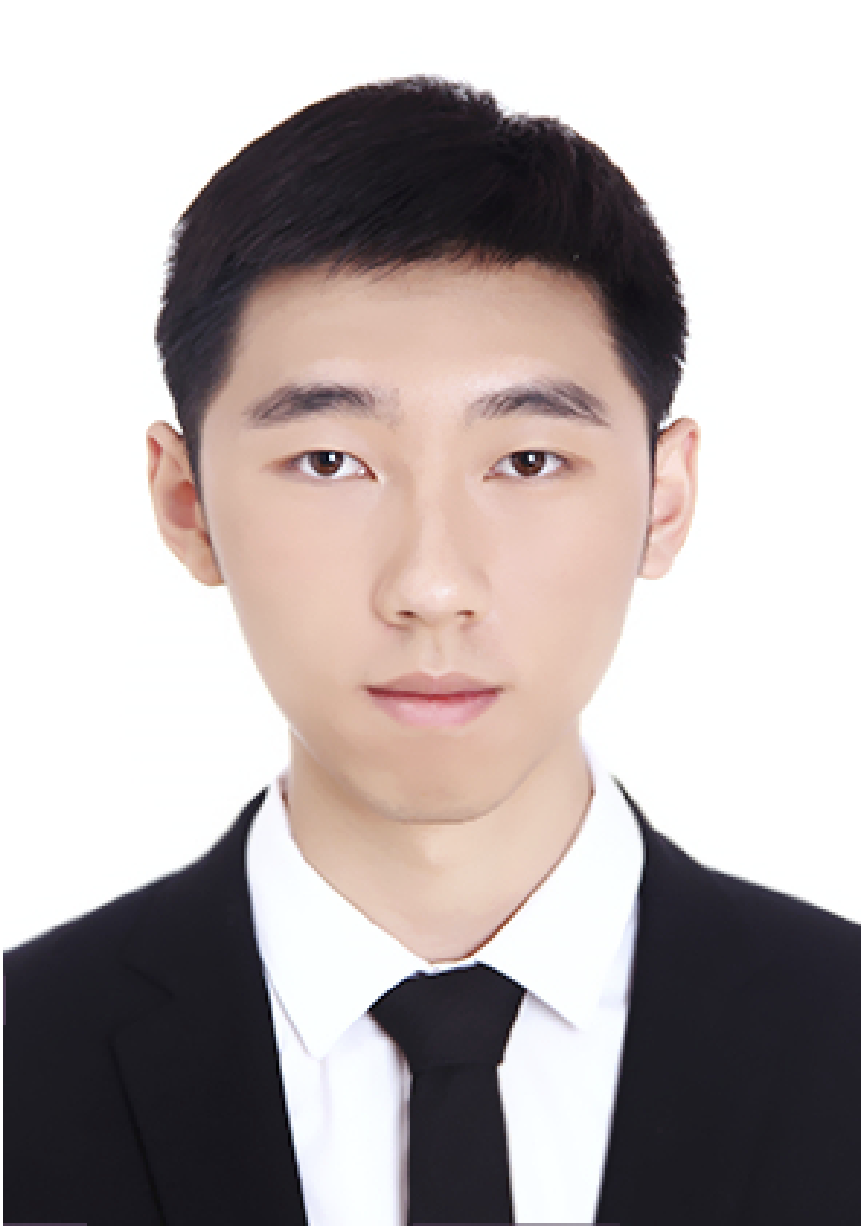}}]{Yihao Liu} received the B.S. degree from University
	of Chinese Academy of Sciences, Beijing, in 2018.
	He is now working towards the Ph.D. degree in Multimedia Laboratory, Shenzhen Institute of Advanced
	Technology, Chinese Academy of Sciences. He is
	supervised by Prof. Yu Qiao and Prof. Chao Dong.
	His research interests include computer vision and
	image/video enhancement.
\end{IEEEbiography}

\begin{IEEEbiography}[{\includegraphics[width=1in,height=1.25in,clip,keepaspectratio]{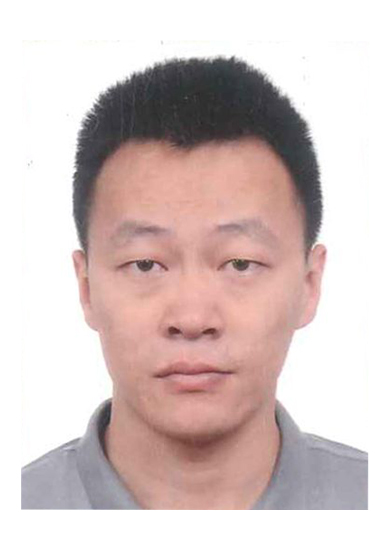}}]{Tong He}    
Tong He received the Ph.D. degree in computer science from the University of Adelaide, Australia, in 2020. He is currently a researcher at Shanghai AI Laboratory. His research interests include computer vision and machine learning.	
\end{IEEEbiography}

\begin{IEEEbiography}[{\includegraphics[width=1in,height=1.25in,clip,keepaspectratio]{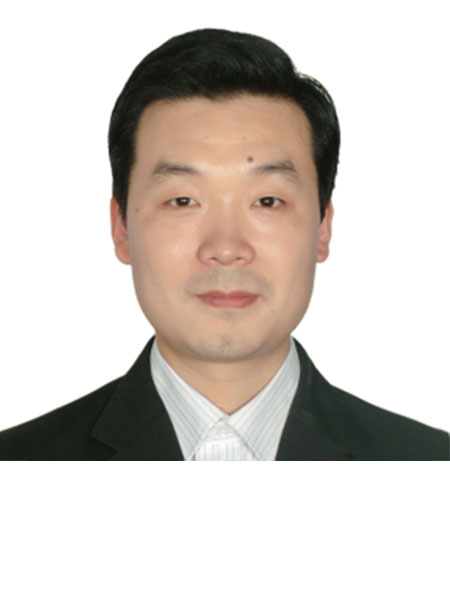}}]{Yu Qiao}
	(SM' 13) received the Ph.D. degree from
	the University of Electro-Communications, Japan,
	in 2006. He was a JSPS Fellow and a Project
	Assistant Professor with the University of Tokyo,
	from 2007 to 2010. He is currently a Professor with
	the Shenzhen Institutes of Advanced Technology,
	Chinese Academy of Sciences. He has authored over
	140 papers in journals and conference including,
	PAMI, IJCV, TIP, ICCV, CVPR, ECCV, and AAAI.
	His research interests include computer vision, deep
	learning, and intelligent robots. He was a recipient
	of the Lu Jiaxi Young Researcher Award from the Chinese Academy of
	Sciences in 2012. He was the first Runner-Up at the ImageNet Large Scale
	Visual Recognition Challenge 2015 in scene recognition and the recipient at
	the ActivityNet Large Scale Activity Recognition Challenge 2016 in video
	classification.
\end{IEEEbiography}

\end{document}